\documentclass[12pt,a4paper]{article}

\usepackage[utf8]{inputenc}
\usepackage[british]{babel}
\usepackage[a4paper,margin=2.5cm]{geometry}
\usepackage{setspace}
\usepackage{microtype}

\usepackage{amsmath,amssymb,amsthm}
\usepackage{algorithm}
\usepackage{algorithmic}
\usepackage{graphicx}
\usepackage{booktabs}
\usepackage{threeparttable}
\usepackage{multirow}
\usepackage{array}
\usepackage{tabularx}
\usepackage{makecell}
\usepackage{adjustbox}
\usepackage{float}
\usepackage[numbers,sort&compress]{natbib}
\usepackage{enumitem}
\usepackage{caption}
\usepackage{subcaption}
\usepackage{xcolor}
\usepackage{listings}
\usepackage{pgfplots}
\usepackage{url}

\usepackage{hyperref}
\hypersetup{
    hidelinks,
    pdftitle={Mini-batch Sampling Strategies for Long-Tailed Image Classification: An Empirical Study on CIFAR-100-LT},
    pdfauthor={Siyu Yuan},
    pdfkeywords={long-tailed recognition, class imbalance, mini-batch sampling, SGD, CIFAR-100-LT}
}

\pgfplotsset{compat=1.18}

\setlist{topsep=6pt, itemsep=3pt, parsep=2pt, partopsep=0pt}

\allowdisplaybreaks[2]

\makeatletter
\renewcommand{\paragraph}{\@startsection{paragraph}{4}{\z@}%
  {2.2ex plus 0.8ex minus 0.4ex}%
  {-0.9em}%
  {\normalfont\normalsize\bfseries}}
\makeatother

\newtheorem{definition}{Definition}

\newtheorem{proposition}{Proposition}

\newcommand{\E}{\mathbb{E}}

\newcommand{\bx}{\mathbf{x}}

\newcommand{\btheta}{\boldsymbol{\theta}}

\begin{document}

\title{Mini-batch Sampling Strategies for\\
       Long-Tailed Image Classification:\\[0.2em]
       An Empirical Study on CIFAR-100-LT}

\author{Siyu Yuan\\
  School of Mathematics, University of Bristol\\
  \texttt{kv23373@alumni.bristol.ac.uk}}

\date{}

\maketitle

\begin{abstract}
Real-world datasets often exhibit long-tailed class distributions, where a few head classes contain a large number of training samples, while a large number of tail classes have only a small number of samples. Each mini-batch composition, determined by the sampling strategy, fundamentally determines which classes contribute to stochastic gradient estimation, thus affecting convergence behaviour and generalisation ability across the entire class spectrum. In this work, we provide a systematic theoretical and empirical comparison of four mini-batch sampling strategies in long-tailed image classification: uniform instance sampling, class-balanced sampling, square root sampling, and progressively balanced sampling. We develop a unified bias-variance framework to characterise the properties of each strategy through its impact on gradient estimation, revealing the fundamental tension between unbiased optimisation of empirical loss and fair representation of rare classes. We evaluate all four strategies under controlled experimental conditions using ResNet-32 on CIFAR-100-LT at three imbalance ratios ($\rho \in \{10, 50, 100\}$), with each strategy sharing the same long-tailed subsets and initialisation within a seed. Our experiments show that progressive sampling improves tail-class accuracy by 25\% relative to the uniform baseline at $\rho = 100$ (13.5\% versus 10.8\%), a gain that is consistent across all three seeds, while its overall accuracy is not distinguishable from that of uniform sampling given the seed-to-seed variation (40.0\% versus 39.7\%). The tail-class gain, not the overall gain, is the robust effect. Notably, at $\rho = 100$ class-balanced sampling degrades performance across every class group---including the tail classes it is designed to help---which we attribute to overfitting caused by extreme oversampling of scarce training samples; at the milder $\rho = 50$ this failure is confined to head and medium classes. These findings suggest that \emph{when} rebalancing is applied during training matters as much as how much rebalancing is applied, and provide guidance for practitioners working with imbalanced datasets.
\end{abstract}


\section{Introduction}
\label{sec:introduction}
Modern machine learning systems are widely used on real-world data. These data rarely conform to the balanced class distribution assumed by standard training procedures. In fields ranging from image analysis to medical diagnosis, from fraud detection to autonomous driving, class distributions follow a characteristic long-tailed pattern, where a few common classes dominate the dataset, while a large number of rare classes contain only a small number of samples \citep{zhang2023dive}. This phenomenon, known as the long-tailed problem, poses a challenge to gradient-based optimisation methods.

Stochastic gradient descent (SGD) and its variants have become a cornerstone of modern machine learning and statistics due to their high training efficiency on large-scale datasets \citep{bottou2018optimization}. The core principle of SGD is to approximate the full gradient using randomly sampled mini-batches, which provides significant computational savings when the dataset contains millions of samples \citep{robbins1951stochastic}. However, this sampling-based approach is highly dependent on the underlying data distribution, resulting in poor performance when classes are imbalanced.

When the training data follows a long-tailed distribution, the mini-batches constructed by standard uniform sampling are dominated by head class samples. Therefore, stochastic gradient estimation tends to minimise the loss of frequent classes, while the gradient signals obtained for rare classes are insufficient. Empirically, this manifests as the model achieving high accuracy on head classes but performing poorly on tail classes, which are often the most important in practice \citep{liu2019large, van2018inaturalist}.

The relationship between mini-batch composition and learning dynamics is central to this challenge. Each mini-batch determines not only the direction of gradient estimation but also its variance. Under uniform sampling from imbalanced data, tail classes may appear in only a small fraction of mini-batches, leading to sporadic and high-variance updates for their associated parameters. This observation motivates the central question of this paper: \textit{how do different mini-batch sampling strategies affect the convergence behaviour and classification performance of SGD on long-tailed data?}

This question admits no simple answer, as different sampling strategies embody different trade-offs. Class-balanced sampling ensures equal representation of all classes but introduces bias into the gradient estimate relative to the true data distribution. Square-root sampling offers an intermediate approach that partially compensates for imbalance without fully equalising class frequencies. Progressively balanced sampling attempts to combine the benefits of different regimes by transitioning from uniform to balanced sampling during training.

\subsection{Contributions}

This paper makes the following contributions:

\begin{enumerate}[leftmargin=*]
    \item We provide a unified mathematical framework for analysing mini-batch sampling strategies in the context of long-tailed learning, characterising the bias and variance properties of each approach.
    
    \item We conduct comprehensive experiments on CIFAR-100-LT across multiple imbalance ratios ($\rho \in \{10, 50, 100\}$), systematically comparing four sampling strategies: uniform instance sampling, class-balanced sampling, square root sampling, and progressively balanced sampling.
    
    \item We analyse the learning dynamics of each strategy, examining convergence behaviour, evolution of class-wise accuracy, and the trade-offs between head and tail class performance.
    
    \item Based on our theoretical and empirical findings, we provide practical recommendations for selecting sampling strategies based on the characteristics of the target dataset and application requirements.
\end{enumerate}

\subsection{Paper Organisation}

The remainder of this paper is organised as follows. Section~\ref{sec:background} reviews the foundations of stochastic gradient descent and defines the long-tailed recognition problem. Section~\ref{sec:methodology} introduces the four sampling strategies under investigation and explains their theoretical properties. Section~\ref{sec:experiments} describes our experimental setup, including the CIFAR-100-LT dataset and training configuration. Section~\ref{sec:results} presents our experimental results and analysis. Section~\ref{sec:discussion} discusses the implications of our findings and their limitations. Section~\ref{sec:conclusion} concludes with a summary and directions for future work.

\section{Background and Related Work}
\label{sec:background}
This section introduces the long-tailed recognition problem and reviews related work on addressing class imbalance. We assume familiarity with stochastic gradient descent; for a comprehensive treatment of its convergence theory, learning rate schedules, and adaptive variants we refer the reader to \citet{robbins1951stochastic}, \citet{bottou2018optimization}, and \citet{bishop2006pattern}.

\subsection{The Long-Tailed Recognition Problem}
\label{subsec:longtail_problem}
In classification problems, the training set $\mathcal{D} = \{(\bx_i, y_i)\}_{i=1}^{n}$ consists of input-label pairs where $y_i \in \{1, \ldots, K\}$ denotes the class label. Let $n_k = |\{i : y_i = k\}|$ denote the number of training samples in class $k$, with $\sum_{k=1}^{K} n_k = n$. Following the standard convention in long-tailed learning \citep{zhang2023deep}, we assume classes are sorted by cardinality in decreasing order: if $i_1 < i_2$, then $n_{i_1} \geq n_{i_2}$, so that $n_1 \geq n_2 \geq \cdots \geq n_K$.

\begin{definition}[Long-tailed distribution \citep{zhang2023deep}]
A class distribution is said to be long-tailed if the class frequencies $\{n_k\}_{k=1}^{K}$ decay rapidly, with a small number of head classes containing the majority of samples and a large number of tail classes containing few samples each.
\end{definition}

The degree of imbalance is commonly quantified by the \textit{imbalance ratio} \citep{zhang2023deep}:
\begin{equation}
    \rho = \frac{n_1}{n_K},
    \label{eq:imbalance_ratio}
\end{equation}
which measures the ratio between the largest and smallest class sizes. Real-world datasets frequently exhibit imbalance ratios ranging from $\rho = 10$ to $\rho > 1000$ \citep{liu2019large}.

A common model for long-tailed distributions is the exponential decay:
\begin{equation}
    n_k = n_1 \cdot \mu^{k-1}, \quad \text{where} \quad \mu = \rho^{-1/(K-1)}.
    \label{eq:exponential_decay}
\end{equation}
This parameterisation ensures that $n_K = n_1 \cdot \mu^{K-1} = n_1 / \rho$, achieving the desired imbalance ratio. The CIFAR-100-LT dataset used in our experiments follows this exponential decay model.

For analysis purposes, it is useful to partition the $K$ classes into three groups based on their frequency: Head classes, Medium classes and Tail classes.
The specific thresholds depend on the dataset and imbalance ratio; we specify our partition for CIFAR-100-LT in Section~\ref{sec:experiments}.

\subsection{The Challenge of Uniform Sampling Under Imbalance}
\label{subsec:uniform_challenge}
Under uniform instance sampling, each training example is selected with probability $1/n$. The probability that a randomly sampled instance belongs to class $k$ is therefore
\begin{equation}
    P(\text{sample from class } k) = \frac{n_k}{n}.
    \label{eq:uniform_class_prob}
\end{equation}

For a mini-batch of size $b$, the expected number of samples from class $k$ is $b \cdot n_k / n$. When class frequencies vary dramatically, this leads to severe under-representation of tail classes. Consider CIFAR-100-LT with $K = 100$ classes and imbalance ratio $\rho = 100$, for which $n = 10{,}847$ (Table~\ref{tab:dataset_stats}). The largest class contains 500 samples and the smallest contains 5, so the mini-batch size $b = 128$ used throughout our experiments (Table~\ref{tab:training_config}) yields, on average:
\begin{itemize}[leftmargin=*]
    \item Expected samples from the largest class: $128 \times 500 / 10{,}847 \approx 5.90$ samples
    \item Expected samples from the smallest class: $128 \times 5 / 10{,}847 \approx 0.059$ samples
\end{itemize}

Equivalently, a given tail class is absent from roughly $94\%$ of mini-batches, whereas head classes appear in almost every batch. The gradient signal for tail classes is therefore both sparse and noisy. This systematic under-representation causes the model to prioritise head class performance at the expense of tail classes.

\subsection{Related Work on Long-Tailed Learning}
\label{subsec:related_work}
The long-tailed recognition problem has attracted substantial research attention, and methods can be broadly categorised into three types.

\textbf{Resampling methods} modify the distribution of training data to achieve more balanced representation across classes. Oversampling replicates samples from minority classes \citep{chawla2002smote}, while undersampling discards samples from majority classes; \citet{buda2018systematic} provide a systematic study of both in convolutional networks. The class-balanced sampling we study in this paper first uniformly and randomly selects classes and then samples instances from those classes \citep{shen2016relay, kang2020decoupling}. These methods directly address sampling bias but may lead to overfitting to the minority class or loss of information from the majority class.

\textbf{Reweighting methods} assign different importance weights to training samples based on class frequency. Cost-sensitive learning assigns higher misclassification costs to minority classes \citep{cui2019class}. Focal loss down-weights well-classified samples to focus learning on hard cases \citep{lin2017focal}. A related line of work leaves the sampler untouched and instead corrects the classifier's outputs using the class priors, either post hoc or through a modified softmax \citep{ren2020balanced, menon2021logit}. These methods modify the loss function rather than the sampling process, providing a complementary perspective on addressing imbalance problems.

\textbf{Transfer learning methods} use knowledge from head classes to improve the performance of the tail classes. This includes feature transfer \citep{liu2019large}, classifier adjustment \citep{saerens2002adjusting}, and meta-learning approaches \citep{wang2017tail}. A further strand decouples representation learning from classifier learning, training the backbone with instance-balanced sampling and only afterwards rebalancing the classifier \citep{kang2020decoupling, zhou2020bbn}. These methods are often used in conjunction with resampling or reweighting strategies.

\paragraph{Relation to prior work.} The four samplers compared here are not new. Instance-balanced, class-balanced, square-root and progressively balanced sampling were assembled and contrasted by \citet{kang2020decoupling} on ImageNet-LT, iNaturalist~2018 and Places-LT, where they served to motivate the decoupling of representation and classifier learning; square-root sampling itself originates in the weakly supervised pretraining of \citet{mahajan2018exploring}. Our contribution is not the strategies but a focused analysis of them. Specifically, we (i) place all four in a single bias--variance decomposition of the mini-batch gradient estimator, from which we derive concrete, falsifiable predictions before looking at the results; (ii) run a controlled comparison on CIFAR-100-LT across three imbalance ratios in which, for a given seed, every strategy sees the identical long-tailed subset and identical initial weights, so that differences are attributable to mini-batch composition alone; (iii) track head, medium and tail accuracy throughout training rather than only at convergence; and (iv) document and diagnose a failure mode that the aggregate numbers in prior work do not isolate---under severe imbalance, class-balanced sampling degrades tail accuracy itself, not merely head accuracy. Our absolute accuracies are consequently not comparable with the large-scale results of \citet{kang2020decoupling}; the object of study is the relative behaviour of the samplers under controlled conditions.

\section{Methodology}
\label{sec:methodology}
This section introduces the four mini-batch sampling strategies and explains their theoretical properties. We focus on characterising each strategy in terms of its effect on gradient estimation, including bias relative to the uniform distribution and variance properties.

\subsection{Problem Setup and Notation}
\label{subsec:setup}
Consider a $K$-class classification problem with training set $\mathcal{D} = \{(\bx_i, y_i)\}_{i=1}^{n}$, where $n_k$ denotes the number of samples in class $k$. We train a neural network with parameters $\btheta$ using cross-entropy loss \citep{bishop2006pattern}:
\begin{equation}
    \mathcal{L}(\btheta) = -\frac{1}{n} \sum_{i=1}^{n} \log p_{y_i}(\bx_i; \btheta),
    \label{eq:cross_entropy}
\end{equation}
where $p_k(\bx; \btheta)$ denotes the predicted probability for class $k$.

Let $\mathcal{L}_k(\btheta)$ denote the average loss over class $k$:
\begin{equation}
    \mathcal{L}_k(\btheta) = -\frac{1}{n_k} \sum_{i: y_i = k} \log p_{y_i}(\bx_i; \btheta).
    \label{eq:class_loss}
\end{equation}

The overall loss Equation~\eqref{eq:cross_entropy} can be written as a weighted combination of class losses:
\begin{equation}
    \mathcal{L}(\btheta) = \sum_{k=1}^{K} \frac{n_k}{n} \mathcal{L}_k(\btheta) = \sum_{k=1}^{K} \pi_k \mathcal{L}_k(\btheta),
    \label{eq:weighted_loss}
\end{equation}
where $\pi_k = n_k / n$ represents the empirical class prior. Different sampling strategies can be understood as implicitly optimising different weightings of the class losses.

\subsection{Sampling Strategies Under Study}
\subsubsection{Strategy 1: Uniform Instance Sampling}
\label{subsec:uniform_sampling}
Uniform instance sampling serves as our baseline \citep{bottou2018optimization}. Each sample is selected independently with probability $1/n$, giving class $k$ a selection probability proportional to its frequency:
\begin{equation}
    P(\text{sample } i) = \frac{1}{n}, \quad P(\text{class } k) = \frac{n_k}{n} = \pi_k.
    \label{eq:uniform_probs}
\end{equation}

\textbf{Gradient estimation.} Let $\hat{\mathcal{L}}$ denote the loss computed on a single mini-batch. The mini-batch gradient is an unbiased estimate of the gradient of Equation~\eqref{eq:cross_entropy}:
\begin{equation}
    \E[\nabla \hat{\mathcal{L}}_{\text{uniform}}] = \nabla \mathcal{L}(\btheta) = \sum_{k=1}^{K} \pi_k \nabla \mathcal{L}_k(\btheta).
    \label{eq:uniform_gradient}
\end{equation}

\textbf{Properties.}
\begin{itemize}[leftmargin=*]

    \item Unbiased: $\E[\nabla\hat{\mathcal{L}}_{\text{uniform}}] = \nabla\mathcal{L}(\btheta)$
    \item Head-dominated gradient: the effective class weight is $w_k = n_k/n$, so at $\rho = 100$ (where $n = 10{,}847$) the largest class contributes $500/10{,}847 \approx 4.6\%$ of the expected gradient signal while the smallest contributes only $5/10{,}847 \approx 0.046\%$---a factor of $100$, exactly the imbalance ratio
    \item Sparse tail updates: the per-class variance contribution is proportional to $w_k^2/n_k = n_k/n^2$ (derived in Section~\ref{sec:bias_variance}), which is \emph{smallest} for tail classes. Their difficulty is therefore not per-update noise but update scarcity: a tail class contributes to the gradient in only about $6\%$ of mini-batches, so its parameters are refined on a correspondingly small number of steps
\end{itemize}

\subsubsection{Strategy 2: Class-Balanced Sampling}
\label{subsec:balanced_sampling}
Class-balanced sampling (CBS) ensures equal representation of all classes regardless of their frequency \citep{shen2016relay, kang2020decoupling}. The sampling proceeds in two stages: first, a class $k$ is selected uniformly at random from $\{1, \ldots, K\}$; then, a sample is drawn uniformly from class $k$.

\begin{equation}
    P(\text{class } k) = \frac{1}{K}, \quad P(\text{sample } i \,|\, \text{class } y_i) = \frac{1}{n_{y_i}}.
    \label{eq:cbs_probs}
\end{equation}

The marginal probability of selecting sample $i$ is:
\begin{equation}
    P(\text{sample } i) = \frac{1}{K} \cdot \frac{1}{n_{y_i}} = \frac{1}{K \cdot n_{y_i}}.
    \label{eq:cbs_marginal}
\end{equation}

Samples from tail classes are selected with higher probability than under uniform sampling, while head class samples are down-weighted.

\textbf{Gradient estimation.} The expected gradient under CBS is:
\begin{equation}
    \E[\nabla \hat{\mathcal{L}}_{\text{CBS}}] = \sum_{k=1}^{K} \frac{1}{K} \nabla \mathcal{L}_k(\btheta) = \nabla \left( \frac{1}{K} \sum_{k=1}^{K} \mathcal{L}_k(\btheta) \right).
    \label{eq:cbs_gradient}
\end{equation}

This corresponds to optimising the \textit{class-balanced loss}---the unweighted average of per-class losses---rather than the instance-weighted loss in Equation~\eqref{eq:cross_entropy}. Relative to uniform sampling, CBS introduces bias:
\begin{equation}
    \text{Bias} = \E[\nabla \hat{\mathcal{L}}_{\text{CBS}}] - \nabla \mathcal{L}(\btheta) = \sum_{k=1}^{K} \left( \frac{1}{K} - \pi_k \right) \nabla \mathcal{L}_k(\btheta).
    \label{eq:cbs_bias}
\end{equation}

\textbf{Properties.}
\begin{itemize}[leftmargin=*]
    \item Biased relative to uniform sampling (but unbiased for class-balanced objective)
    \item Equal expected gradient contribution from each class
    \item Tail classes receive significantly more updates
    \item Risk of overfitting to tail classes due to repeated sampling of limited examples
\end{itemize}

\subsubsection{Strategy 3: Square-Root Sampling}
\label{subsec:sqrt_sampling}
Square-root sampling offers an intermediate approach between uniform and class-balanced sampling. It was introduced by \citet{mahajan2018exploring} for weakly supervised pretraining and adopted as a long-tailed baseline by \citet{kang2020decoupling}. The probability of selecting class $k$ is proportional to the square root of its frequency:
\begin{equation}
    P(\text{class } k) = \frac{\sqrt{n_k}}{\sum_{j=1}^{K} \sqrt{n_j}}.
    \label{eq:sqrt_class_prob}
\end{equation}

Let $Z = \sum_{j=1}^{K} \sqrt{n_j}$ denote the normalisation constant. The marginal probability of selecting sample $i$ from class $y_i$ is:
\begin{equation}
    P(\text{sample } i) = \frac{\sqrt{n_{y_i}}}{Z} \cdot \frac{1}{n_{y_i}} = \frac{1}{Z \sqrt{n_{y_i}}}.
    \label{eq:sqrt_marginal}
\end{equation}

\textbf{Gradient estimation.} The expected gradient is:
\begin{equation}
    \E[\nabla \hat{\mathcal{L}}_{\text{sqrt}}] = \sum_{k=1}^{K} \frac{\sqrt{n_k}}{Z} \nabla \mathcal{L}_k(\btheta).
    \label{eq:sqrt_gradient}
\end{equation}

This corresponds to optimising a loss with class weights proportional to $\sqrt{n_k}$, which partially compensates for class imbalance without fully equalising class contributions.

\textbf{Properties.}
\begin{itemize}[leftmargin=*]
    \item Intermediate bias: $\pi_k \leq \sqrt{n_k}/Z \leq 1/K$ for tail classes, so bias lies between uniform and CBS
    \item Tail classes receive more representation than uniform, less than CBS
    \item Reduces risk of overfitting to tail classes compared to CBS
    \item Effective compromise when some retention of prior information is desired
\end{itemize}

\subsubsection{Strategy 4: Progressively Balanced Sampling}
\label{subsec:progressive_sampling}
Progressively balanced sampling (PBS), proposed by \citet{kang2020decoupling}, recognises that the optimal class weights may change during training. In the early stages of training, it is beneficial to learn general features from rich head class data; later, focusing on tail classes becomes more important. PBS implements this intuition by smoothly transitioning from uniform to class-balanced sampling.

Let $t$ denote the current training epoch and $T$ the total number of epochs. Define the progression parameter:
\begin{equation}
    \lambda(t) = \left(\frac{t}{T-1}\right)^{\gamma},
    \label{eq:progression}
\end{equation}
where $\gamma > 0$ controls the transition speed. The class sampling probability at epoch $t$ is:
\begin{equation}
    P_t(\text{class } k) = (1 - \lambda(t)) \cdot \pi_k + \lambda(t) \cdot \frac{1}{K}.
    \label{eq:pbs_class_prob}
\end{equation}

When $t = 0$, we have $\lambda(0) = 0$ and sampling is purely uniform. When $t = T-1$ (the final epoch), we have $\lambda(T-1) = 1$ and sampling is purely class-balanced. The parameter $\gamma$ controls how quickly the transition occurs:
\begin{itemize}[leftmargin=*]
    \item $\gamma < 1$: rapid early transition (concave schedule)
    \item $\gamma = 1$: linear transition
    \item $\gamma > 1$: delayed transition (convex schedule)
\end{itemize}

\textbf{Properties.}
\begin{itemize}[leftmargin=*]
    \item Time-varying bias: $\|\mathrm{Bias}(t)\| = \lambda(t)\cdot\|\mathrm{Bias}_{\text{CB}}\|$, interpolating from $0$ to full CBS bias
    \item Early training benefits from diverse head class features
    \item Late training focuses on refining tail class performance
    \item Introduces additional hyperparameter $\gamma$ requiring tuning
\end{itemize}

\subsection{Theoretical Analysis: Bias-Variance Trade-offs}
\label{sec:bias_variance}

The four strategies each produce a different stochastic gradient estimator $\hat{g}(\btheta)$. We decompose its mean squared error (MSE) relative to the true empirical gradient $\nabla \mathcal{L}(\btheta)$:
\begin{equation}
    \mathrm{MSE}\bigl[\hat{g}(\btheta)\bigr]
    = \underbrace{\bigl\lVert \mathbb{E}[\hat{g}(\btheta)] - \nabla \mathcal{L}(\btheta) \bigr\rVert^2}_{\text{Bias}^2}
    \;+\; \underbrace{\mathbb{E}\bigl[\bigl\lVert \hat{g}(\btheta) - \mathbb{E}[\hat{g}(\btheta)] \bigr\rVert^2\bigr]}_{\text{Variance}}.
    \label{eq:mse_decomp}
\end{equation}

Each strategy assigns an effective class weight $w_k^{(s)}$ to class $k$, giving expected gradient $\mathbb{E}[\hat{g}_s(\btheta)] = \sum_{k=1}^{K} w_k^{(s)} \nabla \mathcal{L}_k(\btheta)$, with:
\begin{equation}
    w_k^{(s)} =
    \begin{cases}
        \pi_k = n_k / n & \text{Uniform}, \\[4pt]
        1/K & \text{Class-Balanced}, \\[4pt]
        \sqrt{n_k}\,/\,Z & \text{Square-Root}, \\[4pt]
        (1-\lambda)\,\pi_k + \lambda/K & \text{Progressive (at epoch } t\text{)},
    \end{cases}
    \label{eq:class_weights}
\end{equation}
where $Z = \sum_{j=1}^{K}\sqrt{n_j}$ and $\lambda = \lambda(t)$. The bias of strategy $s$ is:
\begin{equation}
    \mathrm{Bias}_s(\btheta) = \sum_{k=1}^{K} \bigl(w_k^{(s)} - \pi_k\bigr)\,\nabla \mathcal{L}_k(\btheta).
    \label{eq:general_bias}
\end{equation}
Uniform sampling achieves zero bias. The bias of class-balanced sampling grows with $\rho$, since $|1/K - \pi_k|$ is largest when the class distribution is highly skewed. The progressive strategy's total weight deviation scales as $\lambda(t)\cdot\lVert\mathrm{Bias}_{\text{CB}}\rVert$, where
$\mathrm{Bias}_{\text{CB}}$ is the full class-balanced bias defined
in Equation~\eqref{eq:cbs_bias}, starting at zero and increasing monotonically.

For the variance term we make two simplifying assumptions, which we state explicitly because the quantitative conclusion below depends on them.

\begin{proposition}[Relative gradient variance under reweighted sampling]
\label{prop:variance}
Assume that (A1) the class composition of a mini-batch of size $b$ is multinomial with class probabilities $w_k^{(s)}$, draws being independent and with replacement; and (A2) the within-class gradient variance is common across classes, $\sigma_k^2 = \sigma^2$ for all $k$. Then the contribution of class $k$ to the variance of $\hat{g}_s(\btheta)$ is proportional to $(w_k^{(s)})^2 \sigma^2 / (b\, n_k)$, and the ratio of the class-balanced to the uniform contribution for a fixed class $k$ is
\begin{equation}
    \frac{\mathrm{Var}_{\text{CB}, k}}{\mathrm{Var}_{\text{uni}, k}} = \frac{(1/K)^2 / n_k}{(n_k/n)^2 / n_k} = \frac{n^2}{K^2 n_k^2}.
    \label{eq:variance_ratio}
\end{equation}
\end{proposition}

\noindent The factor $1/n_k$ reflects the variance of sampling within a class of size $n_k$: a class represented by few distinct examples yields a less diverse, and hence noisier, estimate of its own gradient when sampled repeatedly with replacement. For a tail class with $n_k = 5$ at $\rho = 100$ (where $n = 10{,}847$ and $K = 100$), Equation~\eqref{eq:variance_ratio} evaluates to $10{,}847^2 / (100^2 \times 5^2) \approx 470$, indicating that class-balanced sampling inflates the variance of the tail-class gradient contribution by roughly two and a half orders of magnitude relative to uniform sampling. Assumption (A2) is certainly violated in practice---tail classes are precisely those on which the loss surface is least well determined---but it is conservative here, since a larger $\sigma_k^2$ for small classes would only increase the ratio. This inflation is our proposed explanation for the volatile tail accuracy observed under class-balanced sampling in Section~\ref{sec:results}; we emphasise that Equation~\eqref{eq:variance_ratio} is a statement about gradient noise, not a proof that test accuracy must degrade.

Table~\ref{tab:bias_variance_summary} summarises the trade-off for all strategies; no static strategy simultaneously minimises both bias and variance.

\begin{table}[H]
    \centering
    \caption{Summary of bias-variance properties for each sampling strategy. Relative magnitudes are indicated by Low, Medium, and High labels.}
    \label{tab:bias_variance_summary}
    \begin{threeparttable}
    \begin{tabular}{lcccl}
        \toprule
        \textbf{Strategy} & \textbf{Bias}$^2$ & \multicolumn{2}{c}{\textbf{Gradient Variance}} & \textbf{Description} \\
        \cmidrule(lr){3-4}
         & & Head & Tail & \\
        \midrule
        Uniform        & None   & Low  & Low\tnote{a}  & Unbiased; neglects tail \\
        Class-Balanced & High   & Low  & High           & Equal class weight; overfits tail \\
        Square-Root    & Medium & Low  & Medium         & Intermediate compromise \\
        Progressive    & None\,$\to$\,High & Low & Low\,$\to$\,High & Adaptive schedule \\
        \bottomrule
    \end{tabular}
    \begin{tablenotes}[flushleft]
        \footnotesize
        \item[a] Low per-update variance, but tail classes receive very few updates, causing high estimation error.
    \end{tablenotes}
    \end{threeparttable}
\end{table}

Progressive sampling resolves this dilemma through temporal decomposition. Its MSE at epoch $t$ is:
\begin{equation}
    \mathrm{MSE}_{\text{prog}}(t) = \lambda(t)^2 \cdot \lVert \mathrm{Bias}_{\text{CB}} \rVert^2 + \frac{1}{b}\sum_{k=1}^{K} \frac{\bigl[(1-\lambda)\pi_k + \lambda/K\bigr]^2}{n_k}\,\sigma_k^2,
    \label{eq:mse_progressive}
\end{equation}
where $\sigma_k^2 = \mathrm{Var}_{i:y_i=k}[\nabla \ell(\bx_i; \btheta)]$ denotes the within-class gradient variance for class $k$, with $\ell(\bx_i; \btheta) = -\log p_{y_i}(\bx_i; \btheta)$ being the per-sample cross-entropy loss, $b$ is the mini-batch size, and the bias term scales as $\lambda^2$. During early training ($\lambda \approx 0$), the estimator is nearly unbiased; as $\lambda$ increases, bias grows but the model has already partially converged, attenuating its impact. The learning rate decay at epoch 160 further stabilises the late-stage optimisation by dampening gradient variance when $\lambda \approx 0.80$.

This analysis predicts three experimentally testable outcomes: (1) progressive sampling should match uniform on head classes whilst outperforming it on tail classes; (2) class-balanced sampling should exhibit volatile tail dynamics due to high variance; (3) performance gaps should widen with $\rho$. We verify these predictions in Section~\ref{sec:results}.

\section{Experimental Setup}
\label{sec:experiments}

This section describes the dataset, model architecture, and training configuration used in our experiments. We follow the standard protocol established in the long-tailed recognition literature~\citep{cao2019learning, zhang2023deep} to ensure comparability with prior work.

\subsection{Dataset: CIFAR-100-LT}

We conduct our experiments on CIFAR-100-LT, a long-tailed variant of the CIFAR-100 dataset~\citep{krizhevsky2009learning}. CIFAR-100 contains 60,000 colour images of size $32 \times 32$ pixels across 100 classes, with 500 training images and 100 test images per class. The original dataset is perfectly balanced.

CIFAR-100-LT is constructed by artificially reducing the number of training samples per class according to the exponential decay of Equation~\eqref{eq:exponential_decay}~\citep{cao2019learning}, with $n_1 = 500$ the maximum class size (retained from the original dataset) and $\rho$ the desired imbalance ratio. We experiment with three imbalance ratios:
\begin{itemize}
    \item $\rho = 10$: Mild imbalance, $n_{100} = 49$ samples
    \item $\rho = 50$: Moderate imbalance, $n_{100} = 9$ samples
    \item $\rho = 100$: Severe imbalance, $n_{100} = 5$ samples
\end{itemize}

Table~\ref{tab:dataset_stats} summarises the dataset statistics for each imbalance ratio.

\begin{table}[H]
    \centering
    \caption{CIFAR-100-LT dataset statistics for different imbalance ratios $\rho$.}
    \label{tab:dataset_stats}
    \begin{tabular}{lccc}
        \toprule
        \textbf{Statistic} & $\boldsymbol{\rho = 10}$ & $\boldsymbol{\rho = 50}$ & $\boldsymbol{\rho = 100}$ \\
        \midrule
        Total training samples       & 19,572 & 12,607 & 10,847 \\
        Max class size ($n_1$)       & 500    & 500    & 500    \\
        Min class size ($n_{100}$)   & 49     & 9      & 5      \\
        Median class size            & 158    & 71     & 50     \\
        \bottomrule
    \end{tabular}
\end{table}

The test set remains balanced with 100 images per class (10,000 total), enabling fair evaluation across all classes regardless of their training set frequency.

\paragraph{Class grouping.} To facilitate detailed analysis, we divide the 100 classes into three groups based on their training set frequency, following the many-/medium-/few-shot convention of \citet{liu2019large}:

\begin{itemize}
    \item \textbf{Head} (Many samples): Classes with $n_k > 100$ samples
    \item \textbf{Medium} (Medium samples): Classes with $20 < n_k \leq 100$ samples
    \item \textbf{Tail} (Few samples): Classes with $n_k \leq 20$ samples
\end{itemize}
Note that the number of classes in each group varies with $\rho$. Specifically, for $\rho = 10$ all classes have $n_k \geq 49$, so the tail group is empty. We use a consistent grouping threshold in our experiments to ensure comparability.

\paragraph{Data augmentation.} We apply standard data augmentation to training images: random cropping with 4 pixels of zero padding and random horizontal flipping. Both training and test images are normalised using the CIFAR-100 channel means $(m_R, m_G, m_B) = (0.5071, 0.4867, 0.4408)$ and standard deviations $(s_R, s_G, s_B) = (0.2675, 0.2565, 0.2761)$.

\subsection{Model Architecture}

We use ResNet-32~\citep{he2016deep} as the backbone architecture, which is the standard choice for CIFAR-scale experiments, balancing model capacity with computational efficiency. The architecture consists of:
\begin{itemize}
    \item An initial $3 \times 3$ convolutional layer with 16 filters
    \item Three residual stages with $[16, 32, 64]$ filters and $[5, 5, 5]$ basic blocks per stage
    \item Global average pooling followed by a fully-connected classifier
\end{itemize}

Each basic block contains two $3 \times 3$ convolutional layers with batch normalisation~\citep{ioffe2015batch} and ReLU activations, along with a skip connection. Spatial downsampling by a factor of 2 occurs at the beginning of the second and third stages via strided convolutions. The total parameter count is approximately 0.47 million.

All convolutional layers are initialised using Kaiming (He) normal initialisation~\citep{he2015delving}, batch normalisation parameters are initialised to $\gamma = 1$ and $\beta = 0$, and the final linear layer is initialised from $\mathcal{N}(0, 0.01)$. This initialisation scheme ensures stable gradient flow at the start of training, which is particularly important when using non-uniform sampling strategies that may produce higher-variance gradients.

\subsection{Training Configuration}
\label{subsec:training_config}
All experiments use SGD with momentum as the optimiser, following the standard protocol for CIFAR long-tailed experiments~\citep{cao2019learning}. The training hyperparameters are summarised in Table~\ref{tab:training_config}.

\begin{table}[H]
    \centering
    \caption{Training hyperparameters.}
    \label{tab:training_config}
    \begin{tabular}{ll}
        \toprule
        \textbf{Hyperparameter} & \textbf{Value} \\
        \midrule
        Optimiser           & SGD with Nesterov momentum \\
        Momentum            & 0.9 \\
        Weight decay        & $2 \times 10^{-4}$ \\
        Initial learning rate & 0.1 \\
        Batch size          & 128 \\
        Training epochs     & 200 \\
        Warmup epochs       & 5 (linear) \\
        LR schedule         & Multi-step at epochs $\{160, 180\}$ \\
        LR decay factor     & 0.01 \\
        Gradient clipping   & Max norm 5.0 \\
        \bottomrule
    \end{tabular}
\end{table}

\paragraph{Learning rate schedule.} We employ a multi-step learning rate schedule with linear warmup, as is standard in the long-tailed learning literature~\citep{cao2019learning, zhang2023deep}. During the first 5 epochs, the learning rate increases linearly from $0.02$ to $0.1$ (the base learning rate). The learning rate then remains constant at $0.1$ until epoch 160, at which point it is reduced by a factor of $0.01$ to $0.001$, and again at epoch 180 to $10^{-5}$. This schedule can be expressed as:
\begin{equation}
    \alpha_t = \begin{cases}
        \alpha_0 \cdot \dfrac{t+1}{T_{\mathrm{warm}}} & \text{if } t < T_{\mathrm{warm}} = 5, \\[6pt]
        \alpha_0 \cdot \displaystyle\prod_{m \in \mathcal{M}} \gamma^{\,\mathbf{1}[t \geq m]} & \text{otherwise},
    \end{cases}
    \label{eq:lr_schedule}
\end{equation}
where $\alpha_0 = 0.1$ is the base learning rate, $\mathcal{M} = \{160, 180\}$ is the set of milestones, and $\gamma = 0.01$ is the decay factor. The warmup phase helps stabilise early training when gradient estimates are noisy, which is particularly relevant for the class-balanced and progressive sampling strategies where the effective mini-batch composition differs substantially from the data distribution.

\paragraph{Gradient clipping.} We apply gradient norm clipping with a maximum norm of 5.0 to prevent gradient explosion. This is especially important for class-balanced sampling under severe imbalance ($\rho = 100$), where tail classes with as few as 5 training samples are heavily oversampled, potentially producing large and unstable gradient updates.

\paragraph{Sampling implementation.} For all strategies except uniform, sampling is implemented via \texttt{WeightedRandomSampler} with replacement, where each sample~$i$ with label~$y_i$ is assigned a weight proportional to the desired sampling probability defined in Section~\ref{sec:methodology}. The sampler draws $|\mathcal{D}_{\mathrm{train}}|$ samples per epoch, ensuring that each epoch involves approximately the same number of gradient updates regardless of the strategy. For the progressive strategy, the sampler weights are recomputed at the beginning of each epoch according to the current value of $\lambda(t)$ as defined in Equation~\eqref{eq:progression}. We use \texttt{drop\_last=True} for all training data loaders to ensure consistent batch sizes and stable batch normalisation statistics.

\paragraph{Reproducibility.} Each experiment is repeated with three random seeds $\{42, 123, 456\}$, and we report the mean and standard deviation across runs. To ensure full determinism, we fix all sources of randomness: Python, NumPy, and PyTorch random seeds, as well as CuDNN deterministic mode (\texttt{torch.backends.cudnn.deterministic = True}). DataLoader workers are individually seeded to prevent non-deterministic data loading. The same random seed is used for both dataset construction and model initialisation, ensuring that all four sampling strategies operate on identical long-tailed subsets and start from identical model parameters for each seed.

\paragraph{Evaluation.} We evaluate on the full balanced test set (10,000 images) after every training epoch and report the \emph{best} test accuracy achieved over the 200 epochs, following the standard evaluation protocol. In addition to overall accuracy, we report per-group accuracy (head, medium, tail) to reveal how each sampling strategy affects different frequency groups. All accuracies are computed as the mean per-class accuracy within each group, ensuring that every class contributes equally regardless of test set composition.

\section{Results and Analysis}
\label{sec:results}
This section presents our experimental results comparing four mini-batch sampling strategies across three imbalance ratios. We analyse overall performance, per-group accuracy, convergence dynamics, and the effect of increasing imbalance severity. All results are reported as mean $\pm$ standard deviation over three random seeds.

\subsection{Overall Performance Comparison}

Table~\ref{tab:overall_results} presents the overall test accuracy for each sampling strategy across all three imbalance ratios (see also Figure~\ref{fig:overall_comparison} in Appendix~\ref{app:supp_figures}).

\begin{table}[H]
    \centering
    \caption{Overall test accuracy (\%) on CIFAR-100-LT across different sampling strategies and imbalance ratios. Results are reported as mean $\pm$ std over 3 random seeds.}
    \label{tab:overall_results}
    \begin{tabular}{lccc}
        \toprule
        \textbf{Strategy} & $\boldsymbol{\rho = 10}$ & $\boldsymbol{\rho = 50}$ & $\boldsymbol{\rho = 100}$ \\
        \midrule
        Uniform        & $56.6 \pm 0.1$ & $44.1 \pm 0.3$ & $39.7 \pm 0.4$ \\
        Class-Balanced & $56.0 \pm 0.1$ & $39.9 \pm 0.2$ & $34.0 \pm 0.9$ \\
        Square-Root    & $56.7 \pm 0.1$ & $43.0 \pm 0.2$ & $38.1 \pm 0.2$ \\
        Progressive    & $\mathbf{57.6 \pm 0.2}$ & $\mathbf{44.1 \pm 0.5}$ & $\mathbf{40.0 \pm 0.6}$ \\
        \bottomrule
    \end{tabular}
\end{table}

Several trends emerge from Table~\ref{tab:overall_results}. Progressive sampling attains the highest mean overall accuracy at every imbalance ratio ($57.6\%$, $44.1\%$, $40.0\%$ for $\rho = 10, 50, 100$), but the size of this lead varies greatly: it is clear at $\rho = 10$ ($+1.0$ points over uniform), while at $\rho = 50$ it is a tie to the reported precision ($44.1\%$ for both) and at $\rho = 100$ the $0.3$-point margin is smaller than the seed-to-seed standard deviation of either strategy. We quantify this properly in Section~\ref{sec:paired}, and caution the reader against reading the overall column as establishing a ranking between progressive and uniform sampling. Class-balanced sampling, by contrast, consistently yields the lowest accuracy, with the deficit relative to uniform growing from $0.6$ points at $\rho = 10$ to $5.7$ points at $\rho = 100$; this gap is large relative to the observed seed variability and is the clearest effect in the table. Square-root sampling occupies an intermediate position, consistent with its design as a compromise (Section~\ref{subsec:sqrt_sampling}). The spread across strategies is small at $\rho = 10$ (a $1.6$-point range) but widens to $6.0$ points at $\rho = 100$, indicating that the choice of sampling strategy becomes consequential only under severe imbalance. The per-group analysis in the next section reveals the mechanisms behind these overall trends.

\subsection{Paired Per-Seed Comparison}
\label{sec:paired}

Because all four strategies share the same three seeds---and, within a seed, the same long-tailed subset and the same initial weights (Section~\ref{subsec:training_config})---the runs are naturally paired, and the per-seed differences are more informative than a comparison of means and standard deviations computed independently. Table~\ref{tab:paired} reports the difference between progressive and uniform sampling for each seed.

\begin{table}[H]
    \centering
    \caption{Paired per-seed differences, progressive minus uniform sampling (percentage points). A positive value favours progressive sampling. Per-seed values are taken from Tables~\ref{tab:per_seed_overall}, \ref{tab:per_seed_hmt_rho100} and \ref{tab:per_seed_hmt_rho50}.}
    \label{tab:paired}
    \begin{threeparttable}
    \begin{tabular}{llcccc}
        \toprule
        \textbf{Metric} & $\boldsymbol{\rho}$ & \textbf{Seed 42} & \textbf{Seed 123} & \textbf{Seed 456} & \textbf{Sign agreement} \\
        \midrule
        \multirow{3}{*}{Overall}
        & 10  & $+1.42$ & $+0.75$ & $+0.77$ & 3/3 \\
        & 50  & $+0.54$ & $+0.42$ & $-0.85$ & 2/3 \\
        & 100 & $+0.61$ & $+0.52$ & $-0.27$ & 2/3 \\
        \midrule
        \multirow{2}{*}{Tail}
        & 50  & $+4.00$ & $+2.48$ & $+0.63$ & \textbf{3/3} \\
        & 100 & $+2.13$ & $+3.23$ & $+2.81$ & \textbf{3/3} \\
        \bottomrule
    \end{tabular}
    \end{threeparttable}
\end{table}

The picture is unambiguous once the runs are paired. The overall-accuracy advantage of progressive sampling is \emph{not} robust: at both $\rho = 50$ and $\rho = 100$ one of the three seeds favours uniform sampling, and the mean differences ($+0.04$ and $+0.29$ points) are of the same order as the within-strategy seed variation. The tail-accuracy advantage, on the other hand, holds for every seed at both imbalance ratios at which tail classes exist, with a mean difference of $+2.72$ points at $\rho = 100$ and $+2.37$ points at $\rho = 50$. At $\rho = 100$ this mean difference is about three times the larger of the two per-strategy standard deviations for tail accuracy ($0.9$ points for uniform, $0.7$ for progressive; Table~\ref{tab:full_hmt_rho100}).

With only three seeds we do not report $p$-values; a paired test on $n = 3$ has negligible power, and quoting one would convey more confidence than the design supports. We instead report the raw paired differences and their sign agreement, and throughout the remainder of this paper we treat the tail-class improvement as our substantive finding and the overall-accuracy difference between progressive and uniform sampling as not established.

\subsection{Head vs Medium vs Tail Analysis}
\label{sec:hmt_analysis}

Tables~\ref{tab:hmt_rho100}--\ref{tab:hmt_rho10} present the per-group accuracy breakdown (see also Figure~\ref{fig:hmt_bars} in Appendix~\ref{app:supp_figures} for a visual comparison).

\begin{table}[H]
    \centering
    \caption{Per-group test accuracy (\%) on CIFAR-100-LT with $\rho = 100$. Mean over 3 seeds.}
    \label{tab:hmt_rho100}
    \begin{tabular}{lcccc}
        \toprule
        \textbf{Strategy} & \textbf{Overall} & \textbf{Head} & \textbf{Medium} & \textbf{Tail} \\
        \midrule
        Uniform        & $39.7$ & $\mathbf{66.7}$ & $38.4$ & $10.8$ \\
        Class-Balanced & $34.0$ & $57.7$          & $32.1$ & $9.2$  \\
        Square-Root    & $38.1$ & $63.8$          & $36.5$ & $10.9$ \\
        Progressive    & $\mathbf{40.0}$ & $64.9$ & $\mathbf{38.6}$ & $\mathbf{13.5}$ \\
        \bottomrule
    \end{tabular}
\end{table}

\begin{table}[H]
    \centering
    \caption{Per-group test accuracy (\%) on CIFAR-100-LT with $\rho = 50$. Mean over 3 seeds.}
    \label{tab:hmt_rho50}
    \begin{tabular}{lcccc}
        \toprule
        \textbf{Strategy} & \textbf{Overall} & \textbf{Head} & \textbf{Medium} & \textbf{Tail} \\
        \midrule
        Uniform        & $44.1$ & $\mathbf{65.3}$ & $36.7$ & $13.9$ \\
        Class-Balanced & $39.9$ & $57.7$          & $33.9$ & $14.1$ \\
        Square-Root    & $43.0$ & $62.8$          & $35.6$ & $15.8$ \\
        Progressive    & $\mathbf{44.1}$ & $63.4$ & $\mathbf{37.7}$ & $\mathbf{16.2}$ \\
        \bottomrule
    \end{tabular}
\end{table}

\begin{table}[H]
    \centering
    \caption{Per-group test accuracy (\%) on CIFAR-100-LT with $\rho = 10$. Mean over 3 seeds. The tail group is empty at this imbalance ratio (minimum class size $= 49 > 20$), so tail accuracy is not applicable.}
    \label{tab:hmt_rho10}
    \begin{tabular}{lcccc}
        \toprule
        \textbf{Strategy} & \textbf{Overall} & \textbf{Head} & \textbf{Medium} & \textbf{Tail} \\
        \midrule
        Uniform        & $56.6$ & $63.3$          & $41.7$ & N/A \\
        Class-Balanced & $56.0$ & $62.1$          & $42.4$ & N/A \\
        Square-Root    & $56.7$ & $62.9$          & $42.9$ & N/A \\
        Progressive    & $\mathbf{57.6}$ & $\mathbf{63.7}$ & $\mathbf{44.1}$ & N/A \\
        \bottomrule
    \end{tabular}
\end{table}

The per-group analysis reveals several important findings.

\paragraph{Uniform sampling excels on head classes but neglects the tail.} At $\rho = 100$, uniform sampling achieves the highest head accuracy ($66.7\%$) but a relatively low tail accuracy ($10.8\%$), consistent with the effective class weight $w_k = n_k/n$ concentrating gradient updates on head classes (Section~\ref{subsec:uniform_sampling}). The ratio of head to tail accuracy ($66.7 / 10.8 \approx 6.2$) reflects the extreme imbalance in gradient representation.

\paragraph{Class-balanced sampling suffers across all groups at $\rho = 100$.} A surprising and important result is that under the most severe imbalance, class-balanced sampling does not improve tail accuracy despite being explicitly designed to equalise class representation. At $\rho = 100$, its tail accuracy ($9.2\%$) is actually \emph{lower} than that of uniform sampling ($10.8\%$). This is consistent with the hypothesis that severe overfitting occurs when tail classes with only 5 training samples are heavily oversampled: the model memorises these few examples but fails to generalise to the test set. Simultaneously, the reduced emphasis on head classes degrades head accuracy by $9.0$ points ($57.7\%$ vs $66.7\%$), resulting in the poorest performance across all three groups.

The effect is imbalance-dependent, and we are careful not to overstate it. At $\rho = 50$, where the smallest classes hold 9 rather than 5 samples, class-balanced sampling still loses heavily on head ($-7.6$ points) and medium ($-2.8$ points) classes but \emph{does} deliver a marginal tail improvement over uniform ($14.1\%$ vs $13.9\%$, a difference well within seed noise). The collapse of tail accuracy is therefore specific to the regime in which the oversampled classes are extremely small. This is consistent with the prediction from Section~\ref{sec:bias_variance} that the cost of the biased gradient estimate grows as the class distribution becomes more skewed: the overall deficit relative to uniform grows from $0.6$ points at $\rho = 10$ to $4.2$ at $\rho = 50$ and $5.7$ at $\rho = 100$.

\paragraph{Progressive sampling achieves the best tail performance while maintaining competitive head accuracy.} At $\rho = 100$, progressive sampling achieves $13.5\%$ tail accuracy---a $25\%$ relative improvement over uniform sampling ($10.8\%$)---while sacrificing only $1.8$ points on head classes ($64.9\%$ vs $66.7\%$). As established in Section~\ref{sec:paired}, this tail-class gain is the one effect that holds for every seed at every imbalance ratio where tail classes exist, and it is the finding on which we rest our conclusions. The same pattern holds at $\rho = 50$, where progressive sampling achieves $16.2\%$ tail accuracy versus $13.9\%$ for uniform. The natural interpretation is that the gradual transition from uniform to balanced sampling allows the model to first learn robust feature representations (during the uniform-dominated phase) before shifting focus to tail classes (as $\lambda$ increases); we examine the training dynamics that support this reading in Section~\ref{sec:progressive_analysis}. The corresponding gain in \emph{overall} accuracy at $\rho = 100$ is only $0.3$ points and, as Table~\ref{tab:paired} shows, does not survive a paired per-seed comparison; we therefore do not claim progressive sampling improves overall accuracy over uniform sampling at high $\rho$, only that it substantially rebalances performance towards the tail at no meaningful overall cost.

\paragraph{Mild imbalance reduces strategy differences.} At $\rho = 10$, all strategies achieve similar head accuracies ($62.1$--$63.7\%$) and the tail group is empty since the minimum class size ($n_{100} = 49$) exceeds the tail threshold of 20 samples. The differences are concentrated in the medium group, where progressive sampling still achieves the best accuracy ($44.1\%$ vs $41.7\%$ for uniform). This confirms that sampling strategies have diminishing returns when the data distribution is relatively balanced.
\subsection{Convergence Dynamics}

\begin{figure}[H]
    \centering
    \includegraphics[width=\textwidth]{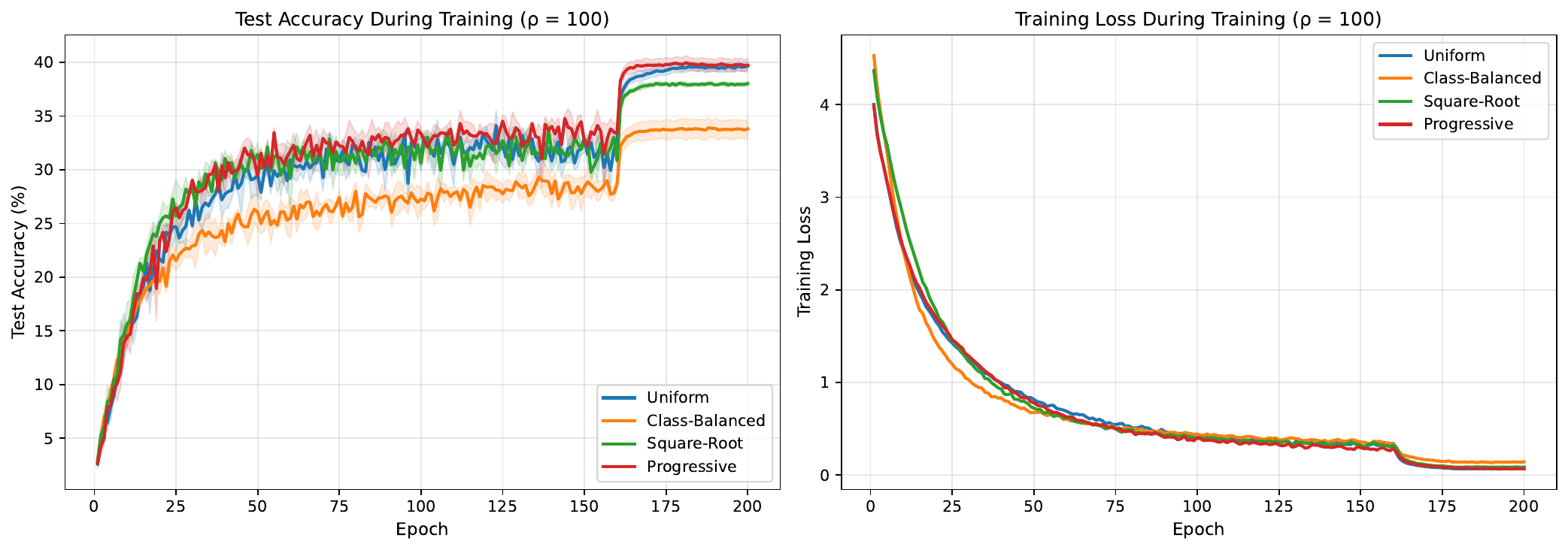}
    \caption{Convergence curves during training for $\rho = 100$. Left: test accuracy. Right: training loss. Shaded regions indicate $\pm 1$ standard deviation over 3 seeds. The learning rate is reduced at epochs 160 and 180 (visible as sharp accuracy jumps). Convergence curves for $\rho = 50$ and $\rho = 10$ are provided in Appendix~\ref{app:convergence_extra}.}
    \label{fig:convergence}
\end{figure}

Figure~\ref{fig:convergence} shows the test accuracy and training loss curves during the training process.

The convergence curves exhibit three distinct phases, controlled by the multi-step learning rate schedule.

\paragraph{Phase 1 (epochs 1-160): Feature learning.} During the main training phase with a learning rate of $0.1$, all strategies show a steady improvement in accuracy. Notably, the relative ordering of strategies is established early in training. By epoch 50, progressive and uniform sampling are clearly ahead of class-balanced sampling. This indicates that low-bias gradient estimates ($\lambda \approx 0$, hence $\mathrm{Bias} \approx 0$) during early training have a lasting impact on the learned feature representations.

\paragraph{Phase 2 (epochs 160--180): First Learning Rate decay.} The reduction in the learning rate at epoch 160 (from $0.1$ to $0.001$) produces a sharp accuracy jump for all strategies. This is the most significant transition in the convergence curves, with accuracy gains of approximately $5$--$8$ percentage points. The magnitude of this jump is largest for progressive sampling, as progressive sampling has transitioned to balanced sampling at this stage ($\lambda(160) = 160/199 \approx 0.80$). The reduced learning rate allows fine-tuning to consolidate the tail class improvements accumulated during the progressive transition.

\paragraph{Phase 3 (epochs 180--200): Convergence.} After the second learning rate decay at epoch 180 (from $0.001$ to $10^{-5}$), accuracy plateaus and all strategies converge to their final values. The narrow confidence band (shaded area) in this phase indicates low variance across seeds.

\paragraph{Training loss behaviour.} The training loss curves (right panels of Figure~\ref{fig:convergence}) show that class-balanced sampling produces a higher training loss during the early epochs, which then decays rapidly. Although all strategies eventually reach near-zero training loss, class-balanced sampling retains a marginally higher final loss than the others. This indicates that the model eventually overfits the training data regardless of sampling strategy. However, despite similar training losses, class-balanced sampling produces significantly lower test accuracy, indicating severe overfitting on the oversampled tail classes.

\subsection{Effect of Imbalance Ratio}
\label{sec:imbalance_effect}

\begin{figure}[H]
    \centering
    \includegraphics[width=\textwidth]{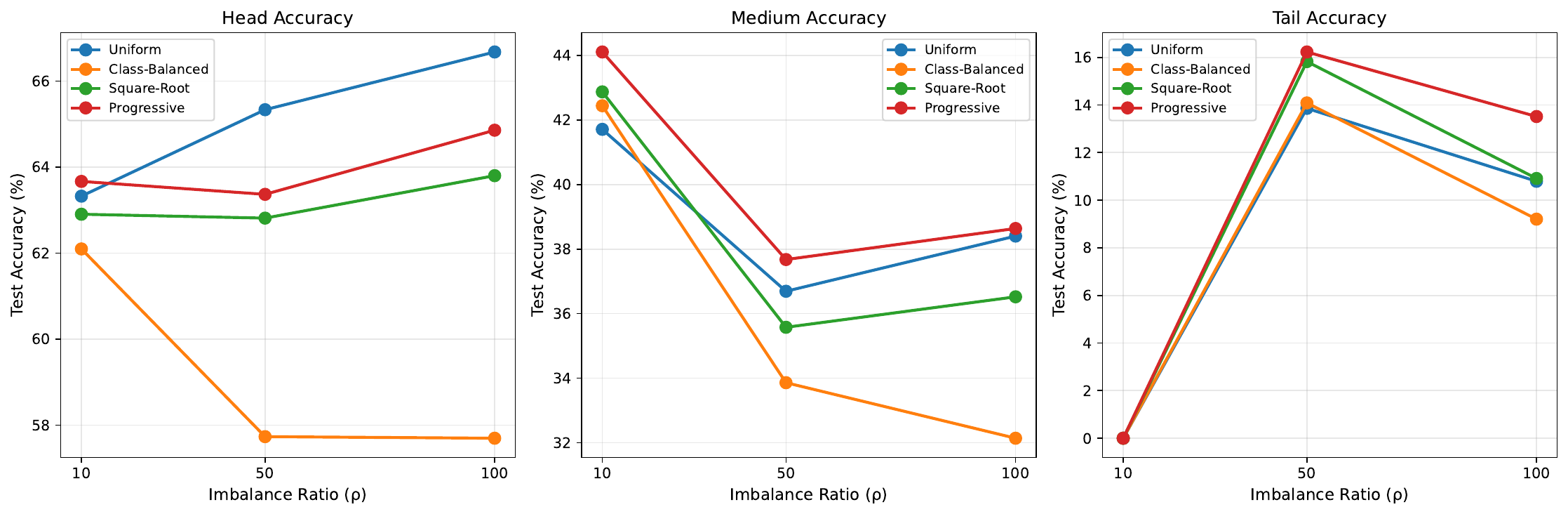}
    \caption{Effect of imbalance ratio on per-group accuracy. Each line tracks a strategy's performance across $\rho \in \{10, 50, 100\}$. In the tail panel, the $\rho = 10$ values are plotted as zero because the tail group is empty at that imbalance ratio; they denote ``not applicable'' and should not be interpreted as measured accuracies.}
    \label{fig:imbalance_effect}
\end{figure}

Figure~\ref{fig:imbalance_effect} examines how the group-by-group performance of each strategy changes as the imbalance ratio increases from $\rho = 10$ to $\rho = 100$.

\paragraph{Head accuracy.} The head accuracy plot reveals a noteworthy pattern: for uniform and progressive sampling, head accuracy increases with imbalance ratio (e.g., uniform head accuracy rises from $63.3\%$ at $\rho = 10$ to $66.7\%$ at $\rho = 100$). This is because at higher $\rho$, a larger proportion of the total training data and gradient updates are used for the head class. Class-balanced sampling shows the opposite trend, with head accuracy declining from $62.1\%$ to $57.7\%$ as $\rho$ increases, because the weight deviation $|1/K - \pi_k|$ grows with $\rho$, making its rebalancing effect more aggressive.

\paragraph{Medium accuracy.} All strategies show a consistent decline in medium accuracy with increasing $\rho$, reflecting the reduced sample sizes for these classes. Progressive sampling maintains the highest medium accuracy across all imbalance ratios, from $44.1\%$ at $\rho = 10$ to $38.6\%$ at $\rho = 100$.

\paragraph{Tail accuracy.} The tail accuracy panel is perhaps the most informative. At $\rho = 10$ the tail group is empty (the minimum class size, 49, exceeds the threshold of 20), so tail accuracy is undefined there. The values plotted at $\rho = 10$ in the tail panel of Figure~\ref{fig:imbalance_effect} are zeros produced by the plotting code and must be read as ``not applicable'' rather than as a measured accuracy of zero; only the $\rho = 50$ and $\rho = 100$ points in that panel carry meaning. At $\rho = 50$ and $\rho = 100$, progressive sampling consistently leads ($16.2\%$ and $13.5\%$ respectively), followed by square-root sampling ($15.8\%$ and $10.9\%$). The decline in tail accuracy from $\rho = 50$ to $\rho = 100$ across all strategies is expected, as it reflects the reduction in minimum class size from 9 to 5 samples. Notably, class-balanced sampling's tail accuracy degrades most dramatically (from $14.1\%$ at $\rho = 50$ to $9.2\%$ at $\rho = 100$), further evidencing the overfitting phenomenon when oversampling extremely small classes.

\subsection{Analysis of Progressive Sampling}
\label{sec:progressive_analysis}

\begin{figure}[H]
    \centering
    \includegraphics[width=\textwidth]{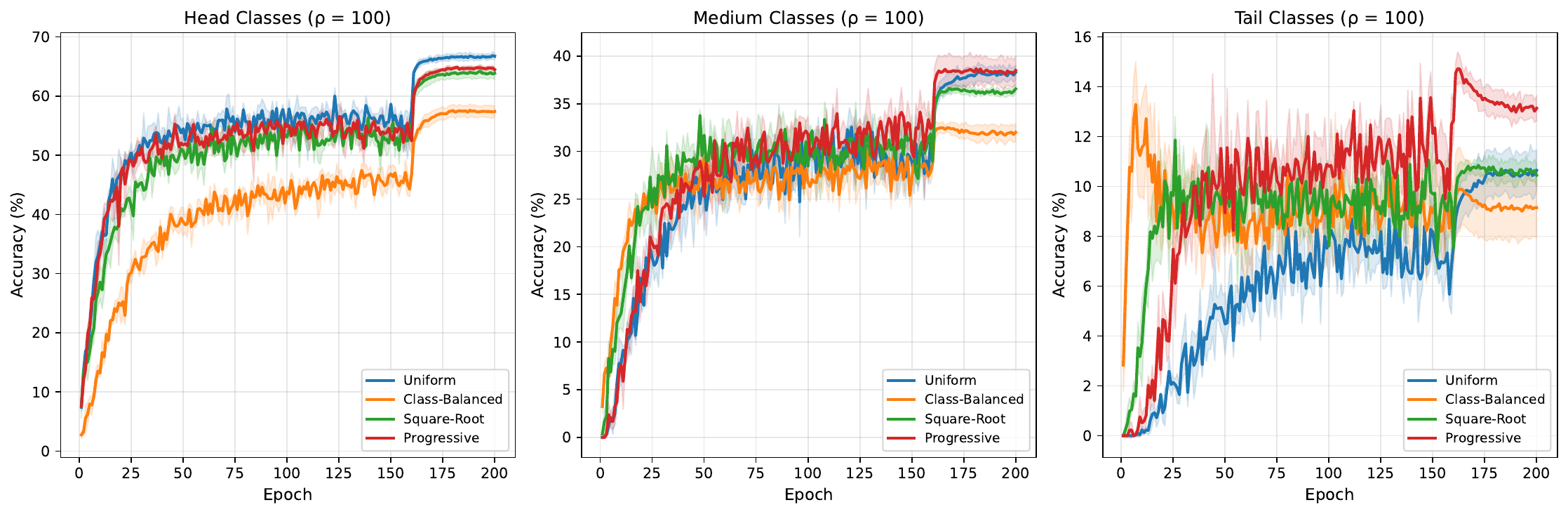}
    \caption{Per-group test accuracy curves during training for $\rho = 100$. Left: head classes. Centre: medium classes. Right: tail classes. Shaded regions indicate $\pm 1$ standard deviation. The corresponding curves for $\rho = 50$ and $\rho = 10$ are provided in Appendix \ref{app:per_group_convergence}.}
    \label{fig:group_curves}
\end{figure}

Progressive sampling's consistent advantage across imbalance ratios and class groups warrants closer investigation. Figure~\ref{fig:group_curves} presents the per-group accuracy evolution during training, which reveals the mechanism behind its success.

Figure~\ref{fig:group_curves} reveals a clear two-phase dynamic: during epochs 1--100, when $\lambda$ is small, progressive sampling tracks the uniform baseline on head and medium classes; as $\lambda$ grows, tail accuracy steadily rises, separating clearly from uniform after the learning rate decay at epoch 160. Crucially, class-balanced sampling shows highly volatile tail accuracy throughout---a signature of overfitting on only 5 training samples per tail class---whereas progressive sampling's steadier trajectory is consistent with the view that introducing rebalancing gradually, once feature representations are already mature, reduces this variance cost.

One caveat deserves emphasis. The learning rate decay at epoch 160 coincides with $\lambda(160) \approx 0.80$, so the reduced step size and the near-balanced sampling regime take effect together, and our design cannot separate their contributions to the tail-class gain that appears at that point. An experiment that varies the milestone location while holding the $\lambda$ schedule fixed (or vice versa) would be needed to attribute the effect; we did not run one, and we therefore describe the coincidence rather than claim a causal interaction.

\section{Discussion}
\label{sec:discussion}
This section synthesises our theoretical framework (Section \ref{sec:methodology}) and experimental findings (Section \ref{sec:results}) to draw broader conclusions about the role of mini-batch sampling in long-tailed learning. We discuss the bias-variance trade-off as observed in practice, provide actionable recommendations for researchers, acknowledge the limitations of our study, and identify directions for future research.

\subsection{The Bias-Variance Trade-off in Practice}

Our experimental results are broadly consistent with the three predictions made in Section~\ref{sec:bias_variance}, with one qualification: progressive sampling does not quite \emph{match} uniform sampling on head classes, giving up $1.8$ points at $\rho = 100$, rather than matching it as predicted. The consistent underperformance of class-balanced sampling reveals a critical limitation of the bias-variance framework when applied na\"ively: the analysis assumes sufficient data within each class to form reliable gradient estimates, an assumption that breaks down when tail classes contain only 5 samples. In this regime, the variance component arising from extreme oversampling dominates, causing the model to memorise the few available tail samples without learning generalisable features---directly evidenced by the volatile tail accuracy curves in Figure~\ref{fig:group_curves}.

\paragraph{Progressive sampling as implicit curriculum learning.} Progressive sampling navigates the bias-variance trade-off through temporal decomposition. By beginning with near-uniform sampling, the model first learns a feature hierarchy driven by the abundant head class data. These features, while biased towards head classes, provide a useful initialisation for all classes since visual categories often share low- and mid-level representations~\citep{zhang2023deep}. As $\lambda$ increases, the model fine-tunes its representations for tail classes, building upon the foundation established during the uniform phase. This two-phase structure can be interpreted as a form of implicit curriculum learning, in the sense of \citet{bengio2009curriculum}, who proposed that training should progress from easy to hard examples. Here, the easy task (learning with abundant head data) precedes the hard task (learning with scarce tail data), mirroring the curriculum principle applied to class imbalance.

\paragraph{Square-root sampling as a static compromise.} Square-root sampling, which assigns class probabilities proportional to $\sqrt{n_k}$, provides a \emph{time-invariant} compromise between the uniform and balanced extremes. Its consistent placement between uniform and class-balanced sampling across all metrics (Tables~\ref{tab:hmt_rho100}--\ref{tab:hmt_rho50}) validates the theoretical prediction that its class weights $w_k = \sqrt{n_k}/Z$ interpolate between $\pi_k$ and $1/K$, and so do its bias-variance properties. However, its inability to match progressive sampling confirms that the \emph{when} of rebalancing matters as much as the \emph{how much}: a fixed intermediate strategy does not capture the stage-dependent requirements of neural network training.

\subsection{Practical Recommendations}

Based on our theoretical analysis and experimental findings, we offer the following practical recommendations for researchers working with long-tailed datasets.

\paragraph{Default to progressive sampling when tail performance matters.} Progressive sampling achieved the best tail-class accuracy at every imbalance ratio at which tail classes exist, consistently across seeds, at a modest cost in head-class accuracy (approximately $1$--$2$ points relative to uniform) and no measurable cost in overall accuracy. Where the application weights rare classes as heavily as common ones---which is the usual reason for caring about long-tailed data in the first place---this is a favourable trade. Where only aggregate accuracy matters, our results give no reason to prefer progressive over uniform sampling, and uniform sampling is simpler. A linear schedule ($\gamma = 1$) from uniform to class-balanced provides a reasonable default, though we tested no other value of $\gamma$ (Section~\ref{sec:discussion}).

\paragraph{Avoid pure class-balanced sampling when the smallest classes are very small.} At $\rho = 100$, where tail classes hold only 5 samples, class-balanced sampling degraded performance across \emph{all} class groups, not just head classes. At $\rho = 50$, with 9 samples in the smallest classes, the same strategy was still clearly worse overall but no longer harmed the tail. We therefore advise against pure class-balanced sampling when $n_{\min}$ is of the order of a handful of examples; if balanced class representation is required in that regime, progressive or square-root sampling should be preferred. We caution that the boundary between these regimes is not resolved by two data points, and that $n_{\min} < 10$ should be read as a rough indication rather than a threshold.

\paragraph{Consider the interaction between the learning rate and sampling schedules.} In our configuration the multi-step learning rate decay at epoch 160 coincides with the late-stage rebalancing of progressive sampling. As noted in Section~\ref{sec:progressive_analysis}, our experiments cannot establish whether this alignment is responsible for part of the observed tail-class gain, since the two were never varied independently. We flag it as a design choice worth attention rather than a validated recommendation: a schedule such as cosine annealing, which lowers the learning rate continuously from the start, would reach the rebalancing phase with a much smaller step size, and whether that helps or hurts is untested. Practitioners adopting progressive sampling may reasonably start from the multi-step schedule used here, with milestones at $80\%$ and $90\%$ of total epochs, while treating the interaction as an open question.


\subsection{Limitations}

Our study has several limitations that should be considered when interpreting the results.

\paragraph{Single architecture and dataset.} All experiments use ResNet-32 on CIFAR-100-LT, a relatively small-scale setting. While CIFAR-100-LT is a standard benchmark for long-tailed learning~\citep{cao2019learning, zhang2023deep}, the absolute performance numbers and relative strategy rankings may differ on larger-scale datasets (e.g., ImageNet-LT~\citep{liu2019large} or iNaturalist~\citep{van2018inaturalist}) or with more powerful architectures (e.g., ResNet-50, Vision Transformers). In particular, larger models with greater capacity may be more or less susceptible to the overfitting effects observed with class-balanced sampling on small tail classes.

\paragraph{Cross-entropy loss only.} We use the standard cross-entropy loss throughout, isolating the effect of sampling strategy from loss function design. In practice, re-sampling is often combined with specialised loss functions such as label-distribution-aware margin loss~\citep{cao2019learning}, focal loss~\citep{lin2017focal}, or class-balanced loss based on effective sample numbers~\citep{cui2019class}. The interaction between sampling strategies and these loss functions is an important dimension that our study does not explore.

\paragraph{Single transition schedule.} For progressive sampling, we use only the linear schedule ($\gamma = 1$). Different transition speeds ($\gamma < 1$ for concave, $\gamma > 1$ for convex) may yield different results, and the optimal $\gamma$ likely depends on the imbalance ratio and dataset characteristics. A systematic study of the transition schedule hyperparameter is beyond the scope of this work.

\paragraph{Synthetic imbalance.} CIFAR-100-LT is constructed by artificially subsampling a balanced dataset, which produces a ``clean'' long-tailed distribution that may not capture the complexity of naturally imbalanced real-world data. Real-world long-tailed datasets often exhibit additional challenges such as visually similar classes, noisy labels, and domain-specific biases that may interact with sampling strategies in ways not captured by our experiments.

\paragraph{Limited evaluation metric.} We report top-1 accuracy as the primary metric. Other metrics such as balanced accuracy (mean per-class accuracy), F1-score, or calibration error may provide complementary perspectives on sampling strategy performance, especially for applications where the cost of misclassification varies across classes.

\paragraph{Model selection on the test set.} We report the best test accuracy achieved over all 200 epochs rather than using a held-out validation set for model selection. While this follows the standard protocol in the long-tailed literature~\citep{cao2019learning, zhang2023deep}, it introduces a mild form of test-set optimism, as the selected checkpoint is the one that maximises performance on the evaluation data. We quantify this directly in Appendix~\ref{app:best_vs_final} by re-deriving every headline number from the final-epoch checkpoint instead: the optimism is between $0.06$ and $0.30$ percentage points, and none of the paper's conclusions depends on the choice of protocol. A dedicated validation split would nonetheless be preferable, and we recommend it for future work.

\subsection{Future Directions}

Our findings have led to several promising research directions.

\paragraph{Adaptive transition schedules.} The fixed linear schedule $\lambda(t) = \left(\frac{t}{T-1}\right)^\gamma$ is a natural starting point, but adaptive approaches that dynamically adjust the transition speed---for example, accelerating $\lambda$ when tail-class validation accuracy plateaus---could further improve performance. Reinforcement learning or meta-learning techniques offer principled ways to learn the optimal schedule.

\paragraph{Combining sampling with complementary methods.} Combining progressive sampling with decoupled training~\citep{kang2020decoupling}, feature-level augmentation for tail classes (e.g., SMOTE~\citep{chawla2002smote}), or specialised loss functions such as label-distribution-aware margin loss~\citep{cao2019learning} represents a natural extension, as these methods address complementary aspects of the long-tailed problem.

\paragraph{Scaling and theoretical extensions.} Validating our findings on larger benchmarks---ImageNet-LT ($\rho = 256$)~\citep{liu2019large} and iNaturalist 2018 ($\rho > 500$)~\citep{van2018inaturalist}---would test the generality of the observed strategy rankings. On the theoretical side, developing a finite-sample analysis of gradient estimation under weighted sampling could provide formal guidelines on when class-balanced sampling becomes harmful, complementing the empirical evidence presented here.

\section{Conclusion}
\label{sec:conclusion}
This work systematically compares four mini-batch sampling strategies---uniform, class-balanced, square-root, and progressively balanced---for long-tailed image classification on CIFAR-100-LT with ResNet-32 across three imbalance ratios ($\rho \in \{10, 50, 100\}$). Through a unified bias-variance framework and controlled experiments, we establish three principal findings.

First, progressive sampling consistently improves tail-class accuracy: it is the best of the four strategies on tail classes at both imbalance ratios where tail classes exist, and the improvement over uniform sampling holds for every random seed (Section~\ref{sec:paired}). Its overall accuracy, by contrast, is indistinguishable from uniform sampling at $\rho = 50$ and $\rho = 100$ once runs are paired by seed, so the strategy is best understood as redistributing accuracy towards rare classes at negligible aggregate cost rather than as a uniform improvement. We read this as a two-phase learning dynamic: early uniform-like sampling builds feature representations from abundant head data, while the gradual transition to balanced sampling redirects learning to underrepresented classes once a representational foundation is in place.

Second, class-balanced sampling proves detrimental under severe imbalance, degrading performance across \emph{all} class groups at $\rho = 100$---including the tail classes it is designed to help. We attribute this to the inflated variance of the tail-class gradient contribution when classes with as few as 5 training samples are heavily oversampled (Proposition~\ref{prop:variance}), a regime that standard bias-variance reasoning glosses over by implicitly assuming sufficient per-class data. At the milder $\rho = 50$ the same strategy still loses substantially on head and medium classes but no longer harms the tail, locating the failure specifically in the small-$n_k$ regime.

Third, the importance of sampling strategy selection grows with imbalance severity: at $\rho = 10$ all strategies perform within a 1.6-point range, whereas at $\rho = 100$ this gap widens to 6.0 points.

The practical implication is that progressive sampling with a linear schedule ($\gamma = 1$) is a sound default for long-tailed training, since it requires no additional hyperparameter tuning, imposes negligible computational overhead, and improves rare-class performance without measurably harming aggregate accuracy---while pure class-balanced sampling should be avoided when the smallest classes are very small. These conclusions are drawn from a single architecture and dataset with three seeds, and we regard confirmation on larger benchmarks as necessary rather than optional. Promising directions for future work include adaptive transition schedules, integration with specialised loss functions such as label-distribution-aware margin loss~\citep{cao2019learning}, and validation on larger benchmarks such as ImageNet-LT~\citep{liu2019large} to test the generality of these conclusions.

\section*{Acknowledgements}
The author thanks Dennis Prangle for supervising this project and for comments on the manuscript.

\section*{Code Availability}
Code to reproduce all experiments, figures and tables in this paper is available at
\url{https://github.com/SiyuYuan0117/longtail-sampling}. The repository contains the
CIFAR-100-LT construction, the four samplers of Section~\ref{sec:methodology}, the training
pipeline described in Section~\ref{subsec:training_config}, and the per-epoch logs underlying
the reported results.

\newpage
\bibliographystyle{plainnat}
\bibliography{references}
\newpage
\appendix

\section*{Appendices}

\definecolor{codebg}{rgb}{0.97, 0.97, 0.97}
\definecolor{codegreen}{rgb}{0.0, 0.5, 0.0}
\definecolor{codegray}{rgb}{0.5, 0.5, 0.5}
\definecolor{codepurple}{rgb}{0.5, 0.0, 0.5}

\lstdefinestyle{pythonstyle}{
    language=Python,
    backgroundcolor=\color{codebg},
    basicstyle=\ttfamily\scriptsize,
    keywordstyle=\color{codepurple}\bfseries,
    stringstyle=\color{codegreen},
    commentstyle=\color{codegray}\itshape,
    numberstyle=\tiny\color{codegray},
    numbers=left,
    numbersep=5pt,
    breaklines=true,
    breakatwhitespace=false,
    tabsize=4,
    showstringspaces=false,
    frame=single,
    rulecolor=\color{codegray!30},
    captionpos=b,
    aboveskip=8pt,
    belowskip=8pt,
    xleftmargin=15pt,
    framexleftmargin=15pt,
}
\lstset{style=pythonstyle}

\section{Implementation Details}
\label{app:code}

This appendix provides the core implementation of the sampling strategies and training pipeline used in our experiments. The complete source code is available in the repository linked in the Code Availability statement. All experiments are implemented in Python 3.10 with PyTorch 2.1.

\subsection{Sampling Strategy Implementation}
\label{app:sampler_code}

Listing~\ref{lst:sampler} presents the implementation of all four sampling strategies. Each strategy is realised via PyTorch's \texttt{WeightedRandomSampler}, which draws samples with replacement according to per-instance weights. Uniform sampling uses the default \texttt{DataLoader} shuffle instead.

\begin{lstlisting}[caption={Implementation of the four sampling strategies.}, label=lst:sampler]
def create_sampler(strategy, labels, class_counts, epoch=0, total_epochs=200):
    """
    Create a WeightedRandomSampler for the given strategy.
    
    Args:
        strategy: 'uniform', 'class_balanced', 'square_root', 'progressive'
        labels: list of class labels for each sample
        class_counts: list of sample counts per class
        epoch: current epoch (for progressive)
        total_epochs: total training epochs
    
    Returns:
        WeightedRandomSampler or None (for uniform)
    """
    num_classes = len(class_counts)
    n_total = sum(class_counts)
    
    if strategy == 'uniform':
        return None  # Use DataLoader shuffle
    
    elif strategy == 'class_balanced':
        # P(class k) = 1/K, P(sample i) = 1 / (K * n_{y_i})
        weights = []
        for label in labels:
            w = 1.0 / (num_classes * max(class_counts[label], 1))
            weights.append(w)
    
    elif strategy == 'square_root':
        # P(class k) proportional to sqrt(n_k)
        sqrt_counts = [np.sqrt(max(c, 1)) for c in class_counts]
        Z = sum(sqrt_counts)
        weights = []
        for label in labels:
            w = np.sqrt(max(class_counts[label], 1)) \
                / (Z * max(class_counts[label], 1))
            weights.append(w)
    
    elif strategy == 'progressive':
        # Linear interpolation: uniform -> class-balanced
        if total_epochs > 1:
            lam = min(epoch / (total_epochs - 1), 1.0)
        else:
            lam = 1.0
        
        # P_eff(k) = (1 - lam) * (n_k / n) + lam * (1 / K)
        class_probs = []
        for k in range(num_classes):
            p_uniform = class_counts[k] / n_total
            p_balanced = 1.0 / num_classes
            p_eff = (1.0 - lam) * p_uniform + lam * p_balanced
            class_probs.append(p_eff)
        
        weights = []
        for label in labels:
            w = class_probs[label] / max(class_counts[label], 1)
            weights.append(w)
    
    weights_tensor = torch.DoubleTensor(weights)
    return WeightedRandomSampler(
        weights_tensor, num_samples=len(weights_tensor), replacement=True
    )
\end{lstlisting}

\subsection{Learning Rate Schedule}
\label{app:lr_code}

Listing~\ref{lst:lr} implements the multi-step learning rate schedule with linear warmup described in Section~\ref{subsec:training_config}.

\begin{lstlisting}[caption={Multi-step learning rate schedule with linear warmup.}, label=lst:lr]
def get_lr(epoch, config):
    """
    Multi-step LR with linear warmup.
    Warmup: linearly increase from 0 to base_lr over warmup_epochs.
    Multi-step: decay by lr_gamma at each milestone.
    """
    base_lr = config['lr']           # 0.1
    warmup_epochs = config['warmup_epochs']  # 5
    
    if epoch < warmup_epochs:
        return base_lr * (epoch + 1) / warmup_epochs
    
    lr = base_lr
    for milestone in config['lr_milestones']:  # [160, 180]
        if epoch >= milestone:
            lr *= config['lr_gamma']           # 0.01
    
    return lr
\end{lstlisting}

\subsection{Training Loop}
\label{app:train_code}

Listing~\ref{lst:train} shows the core training loop for a single epoch, including gradient clipping.

\begin{lstlisting}[caption={Training loop for one epoch with gradient clipping.}, label=lst:train]
def train_one_epoch(model, loader, criterion, optimizer, config):
    """Train for one epoch with gradient clipping."""
    model.train()
    total_loss = 0.0
    correct = 0
    total = 0
    
    for inputs, targets in loader:
        inputs, targets = inputs.to(DEVICE), targets.to(DEVICE)
        
        optimizer.zero_grad()
        outputs = model(inputs)
        loss = criterion(outputs, targets)
        loss.backward()
        
        # Gradient clipping for stability
        if config['grad_clip'] > 0:
            nn.utils.clip_grad_norm_(
                model.parameters(), config['grad_clip']
            )
        
        optimizer.step()
        
        total_loss += loss.item() * inputs.size(0)
        _, predicted = outputs.max(1)
        total += targets.size(0)
        correct += predicted.eq(targets).sum().item()
    
    return total_loss / total, 100.0 * correct / total
\end{lstlisting}

\subsection{Evaluation with Per-Group Accuracy}
\label{app:eval_code}

Listing~\ref{lst:eval} shows the evaluation function that computes overall and per-group (head/medium/tail) accuracies.

\begin{lstlisting}[caption={Evaluation function with head/medium/tail group accuracy computation.}, label=lst:eval]
def evaluate(model, loader, criterion, class_counts=None, num_classes=100):
    """Evaluate model; return overall and per-group accuracy."""
    model.eval()
    class_correct = np.zeros(num_classes)
    class_total = np.zeros(num_classes)
    
    with torch.no_grad():
        for inputs, targets in loader:
            inputs, targets = inputs.to(DEVICE), targets.to(DEVICE)
            outputs = model(inputs)
            _, predicted = outputs.max(1)
            for i in range(targets.size(0)):
                label = targets[i].item()
                class_total[label] += 1
                if predicted[i].item() == label:
                    class_correct[label] += 1
    
    per_class_acc = np.where(
        class_total > 0,
        100.0 * class_correct / class_total, 0.0
    )
    
    # Group by training frequency
    head_idx   = [i for i in range(num_classes) if class_counts[i] > 100]
    medium_idx = [i for i in range(num_classes) if 20 < class_counts[i] <= 100]
    tail_idx   = [i for i in range(num_classes) if class_counts[i] <= 20]
    
    head_acc   = np.mean([per_class_acc[i] for i in head_idx])   if head_idx   else 0.0
    medium_acc = np.mean([per_class_acc[i] for i in medium_idx]) if medium_idx else 0.0
    tail_acc   = np.mean([per_class_acc[i] for i in tail_idx])   if tail_idx   else 0.0
    
    return {
        'overall_acc': 100.0 * np.sum(class_correct) / np.sum(class_total),
        'head_acc': head_acc, 'medium_acc': medium_acc, 'tail_acc': tail_acc,
    }
\end{lstlisting}

\section{Complete Experimental Results}
\label{app:full_results}

This section provides the full per-seed results that underlie the summary statistics reported in Section~\ref{sec:results}.

\subsection{Per-Seed Overall Accuracy}

Table~\ref{tab:per_seed_overall} reports the best test accuracy for each individual run, along with the mean and standard deviation across the three seeds.

\begin{table}[H]
    \centering
    \caption{Overall test accuracy (\%) for each seed. Bold indicates the best strategy per $\rho$.}
    \label{tab:per_seed_overall}
    \begin{tabular}{llcccc}
        \toprule
        $\boldsymbol{\rho}$ & \textbf{Strategy} & \textbf{Seed 42} & \textbf{Seed 123} & \textbf{Seed 456} & \textbf{Mean $\pm$ Std} \\
        \midrule
        \multirow{4}{*}{10}
        & Uniform        & 56.48 & 56.80 & 56.59 & $56.6 \pm 0.1$ \\
        & Class-Balanced & 55.94 & 55.90 & 56.17 & $56.0 \pm 0.1$ \\
        & Square-Root    & 56.59 & 56.68 & 56.81 & $56.7 \pm 0.1$ \\
        & Progressive    & \textbf{57.90} & \textbf{57.55} & \textbf{57.36} & $\mathbf{57.6 \pm 0.2}$ \\
        \midrule
        \multirow{4}{*}{50}
        & Uniform        & 44.19 & 43.66 & 44.44 & $44.1 \pm 0.3$ \\
        & Class-Balanced & 40.18 & 39.78 & 39.71 & $39.9 \pm 0.2$ \\
        & Square-Root    & 43.23 & 43.07 & 42.67 & $43.0 \pm 0.2$ \\
        & Progressive    & \textbf{44.73} & \textbf{44.08} & \textbf{43.59} & $\mathbf{44.1 \pm 0.5}$ \\
        \midrule
        \multirow{4}{*}{100}
        & Uniform        & 40.27 & 39.30 & 39.65 & $39.7 \pm 0.4$ \\
        & Class-Balanced & 34.13 & 32.82 & 34.97 & $34.0 \pm 0.9$ \\
        & Square-Root    & 37.84 & 38.33 & 38.22 & $38.1 \pm 0.2$ \\
        & Progressive    & \textbf{40.88} & \textbf{39.82} & \textbf{39.38} & $\mathbf{40.0 \pm 0.6}$ \\
        \bottomrule
    \end{tabular}
\end{table}

\subsection{Detailed Head/Medium/Tail Accuracy}

Tables~\ref{tab:full_hmt_rho100}--\ref{tab:full_hmt_rho10} provide the per-group accuracy with standard deviations for each imbalance ratio.

\begin{table}[H]
    \centering
    \caption{Detailed per-group accuracy (\%) for $\rho = 100$. Mean $\pm$ std over 3 seeds.}
    \label{tab:full_hmt_rho100}
    \begin{tabular}{lcccc}
        \toprule
        \textbf{Strategy} & \textbf{Overall} & \textbf{Head} & \textbf{Medium} & \textbf{Tail} \\
        \midrule
        Uniform        & $39.7 \pm 0.4$ & $\mathbf{66.7 \pm 0.6}$ & $38.4 \pm 0.8$ & $10.8 \pm 0.9$ \\
        Class-Balanced & $34.0 \pm 0.9$ & $57.7 \pm 1.1$ & $32.1 \pm 0.8$ & $9.2 \pm 1.2$ \\
        Square-Root    & $38.1 \pm 0.2$ & $63.8 \pm 0.8$ & $36.5 \pm 0.4$ & $10.9 \pm 0.4$ \\
        Progressive    & $\mathbf{40.0 \pm 0.6}$ & $64.9 \pm 0.1$ & $\mathbf{38.6 \pm 1.5}$ & $\mathbf{13.5 \pm 0.7}$ \\
        \bottomrule
    \end{tabular}
\end{table}

\begin{table}[H]
    \centering
    \caption{Detailed per-group accuracy (\%) for $\rho = 50$. Mean $\pm$ std over 3 seeds.}
    \label{tab:full_hmt_rho50}
    \begin{tabular}{lcccc}
        \toprule
        \textbf{Strategy} & \textbf{Overall} & \textbf{Head} & \textbf{Medium} & \textbf{Tail} \\
        \midrule
        Uniform        & $44.1 \pm 0.3$ & $\mathbf{65.3 \pm 0.4}$ & $36.7 \pm 0.8$ & $13.9 \pm 0.0$ \\
        Class-Balanced & $39.9 \pm 0.2$ & $57.7 \pm 0.2$ & $33.9 \pm 0.0$ & $14.1 \pm 0.6$ \\
        Square-Root    & $43.0 \pm 0.2$ & $62.8 \pm 0.8$ & $35.6 \pm 0.8$ & $15.8 \pm 0.7$ \\
        Progressive    & $\mathbf{44.1 \pm 0.5}$ & $63.4 \pm 0.3$ & $\mathbf{37.7 \pm 0.7}$ & $\mathbf{16.2 \pm 1.4}$ \\
        \bottomrule
    \end{tabular}
\end{table}

\begin{table}[H]
    \centering
    \caption{Detailed per-group accuracy (\%) for $\rho = 10$. Mean $\pm$ std over 3 seeds. The tail group is empty (all classes have $n_k \geq 49 > 20$).}
    \label{tab:full_hmt_rho10}
    \begin{tabular}{lcccc}
        \toprule
        \textbf{Strategy} & \textbf{Overall} & \textbf{Head} & \textbf{Medium} & \textbf{Tail} \\
        \midrule
        Uniform        & $56.6 \pm 0.1$ & $63.3 \pm 0.1$ & $41.7 \pm 0.5$ & N/A \\
        Class-Balanced & $56.0 \pm 0.1$ & $62.1 \pm 0.4$ & $42.4 \pm 0.6$ & N/A \\
        Square-Root    & $56.7 \pm 0.1$ & $62.9 \pm 0.4$ & $42.9 \pm 1.2$ & N/A \\
        Progressive    & $\mathbf{57.6 \pm 0.2}$ & $\mathbf{63.7 \pm 0.2}$ & $\mathbf{44.1 \pm 0.8}$ & N/A \\
        \bottomrule
    \end{tabular}
\end{table}

\subsection{Per-Seed Head/Medium/Tail Breakdown}
\label{app:per_seed_hmt}

Tables~\ref{tab:per_seed_hmt_rho100}--\ref{tab:per_seed_hmt_rho10} provide the full per-seed breakdown of head, medium, and tail accuracy for each imbalance ratio, complementing the summary statistics in Section~\ref{app:full_results}.

\begin{table}[H]
    \centering
    \caption{Per-seed head/medium/tail accuracy (\%) for $\rho = 100$.}
    \label{tab:per_seed_hmt_rho100}
    \resizebox{\textwidth}{!}{%
    \begin{tabular}{llccccc}
        \toprule
        \textbf{Strategy} & \textbf{Group} & \textbf{Seed 42} & \textbf{Seed 123} & \textbf{Seed 456} & \textbf{Mean} & \textbf{Std} \\
        \midrule
        \multirow{3}{*}{Uniform}
        & Head   & 66.14 & 66.40 & 67.49 & 66.7 & 0.6 \\
        & Medium & 39.44 & 37.47 & 38.29 & 38.4 & 0.8 \\
        & Tail   & 11.97 & 10.71 &  9.71 & 10.8 & 0.9 \\
        \midrule
        \multirow{3}{*}{Class-Balanced}
        & Head   & 57.66 & 56.37 & 59.06 & 57.7 & 1.1 \\
        & Medium & 31.56 & 31.62 & 33.24 & 32.1 & 0.8 \\
        & Tail   & 10.39 &  7.55 &  9.68 &  9.2 & 1.2 \\
        \midrule
        \multirow{3}{*}{Square-Root}
        & Head   & 62.69 & 64.03 & 64.69 & 63.8 & 0.8 \\
        & Medium & 36.35 & 37.06 & 36.15 & 36.5 & 0.4 \\
        & Tail   & 11.42 & 10.71 & 10.61 & 10.9 & 0.4 \\
        \midrule
        \multirow{3}{*}{Progressive}
        & Head   & 64.74 & 64.83 & 65.00 & 64.9 & 0.1 \\
        & Medium & 40.74 & 37.68 & 37.50 & 38.6 & 1.5 \\
        & Tail   & 14.10 & 13.94 & 12.52 & 13.5 & 0.7 \\
        \bottomrule
    \end{tabular}%
    }
\end{table}

\begin{table}[H]
    \centering
    \caption{Per-seed head/medium/tail accuracy (\%) for $\rho = 50$.}
    \label{tab:per_seed_hmt_rho50}
    \resizebox{\textwidth}{!}{%
    \begin{tabular}{llccccc}
        \toprule
        \textbf{Strategy} & \textbf{Group} & \textbf{Seed 42} & \textbf{Seed 123} & \textbf{Seed 456} & \textbf{Mean} & \textbf{Std} \\
        \midrule
        \multirow{3}{*}{Uniform}
        & Head   & 64.85 & 65.34 & 65.80 & 65.3 & 0.4 \\
        & Medium & 37.40 & 35.58 & 37.10 & 36.7 & 0.8 \\
        & Tail   & 13.89 & 13.89 & 13.79 & 13.9 & 0.0 \\
        \midrule
        \multirow{3}{*}{Class-Balanced}
        & Head   & 58.05 & 57.46 & 57.68 & 57.7 & 0.2 \\
        & Medium & 33.90 & 33.85 & 33.83 & 33.9 & 0.0 \\
        & Tail   & 14.84 & 14.11 & 13.32 & 14.1 & 0.6 \\
        \midrule
        \multirow{3}{*}{Square-Root}
        & Head   & 61.98 & 63.98 & 62.49 & 62.8 & 0.8 \\
        & Medium & 36.65 & 35.00 & 35.08 & 35.6 & 0.8 \\
        & Tail   & 16.63 & 14.95 & 15.89 & 15.8 & 0.7 \\
        \midrule
        \multirow{3}{*}{Progressive}
        & Head   & 63.05 & 63.71 & 63.34 & 63.4 & 0.3 \\
        & Medium & 38.70 & 37.12 & 37.20 & 37.7 & 0.7 \\
        & Tail   & 17.89 & 16.37 & 14.42 & 16.2 & 1.4 \\
        \bottomrule
    \end{tabular}%
    }
\end{table}

\begin{table}[H]
    \centering
    \caption{Per-seed head/medium accuracy (\%) for $\rho = 10$. Tail group is empty at this imbalance ratio.}
    \label{tab:per_seed_hmt_rho10}
    \resizebox{\textwidth}{!}{%
    \begin{tabular}{llccccc}
        \toprule
        \textbf{Strategy} & \textbf{Group} & \textbf{Seed 42} & \textbf{Seed 123} & \textbf{Seed 456} & \textbf{Mean} & \textbf{Std} \\
        \midrule
        \multirow{2}{*}{Uniform}
        & Head   & 63.38 & 63.35 & 63.25 & 63.3 & 0.1 \\
        & Medium & 41.13 & 42.23 & 41.77 & 41.7 & 0.5 \\
        \midrule
        \multirow{2}{*}{Class-Balanced}
        & Head   & 61.65 & 62.01 & 62.62 & 62.1 & 0.4 \\
        & Medium & 43.23 & 42.29 & 41.81 & 42.4 & 0.6 \\
        \midrule
        \multirow{2}{*}{Square-Root}
        & Head   & 63.46 & 62.74 & 62.51 & 62.9 & 0.4 \\
        & Medium & 41.29 & 43.19 & 44.13 & 42.9 & 1.2 \\
        \midrule
        \multirow{2}{*}{Progressive}
        & Head   & 63.75 & 63.42 & 63.83 & 63.7 & 0.2 \\
        & Medium & 44.87 & 44.48 & 42.97 & 44.1 & 0.8 \\
        \bottomrule
    \end{tabular}%
    }
\end{table}

\section{Best-Epoch versus Final-Epoch Results}
\label{app:best_vs_final}

Throughout the paper we follow the common protocol of reporting the best test accuracy observed over the 200 training epochs. Because the selection is made on the evaluation set, this introduces a degree of test-set optimism, which we quantify here by recomputing every headline quantity from the final-epoch (epoch 200) checkpoint, a protocol that involves no selection whatsoever.

Table~\ref{tab:best_vs_final} reports both. The optimism is small in absolute terms: the gap ranges from $0.06$ to $0.30$ percentage points with a mean of $0.16$, an order of magnitude smaller than the differences between strategies that we interpret in Section~\ref{sec:results}. The reason is visible in the convergence curves of Figure~\ref{fig:convergence}: after the second learning rate decay at epoch 180 the accuracy plateaus, so the best epoch falls late in training (between epochs 167 and 200 across all 36 runs) and differs little from the last.

\begin{table}[H]
    \centering
    \caption{Best-epoch versus final-epoch (epoch 200) overall test accuracy (\%). ``Best'' is the protocol used throughout the paper; ``Final'' uses the last checkpoint and involves no selection on the test set. Mean $\pm$ std over 3 seeds. The gap column is Best $-$ Final.}
    \label{tab:best_vs_final}
    \begin{tabular}{llccc}
        \toprule
        $\boldsymbol{\rho}$ & \textbf{Strategy} & \textbf{Best} & \textbf{Final} & \textbf{Gap} \\
        \midrule
        \multirow{4}{*}{10}
        & Uniform        & $56.6 \pm 0.1$ & $56.4 \pm 0.1$ & $0.20$ \\
        & Class-Balanced & $56.0 \pm 0.1$ & $55.9 \pm 0.2$ & $0.09$ \\
        & Square-Root    & $56.7 \pm 0.1$ & $56.5 \pm 0.1$ & $0.20$ \\
        & Progressive    & $57.6 \pm 0.2$ & $57.5 \pm 0.2$ & $0.13$ \\
        \midrule
        \multirow{4}{*}{50}
        & Uniform        & $44.1 \pm 0.3$ & $44.0 \pm 0.3$ & $0.13$ \\
        & Class-Balanced & $39.9 \pm 0.2$ & $39.8 \pm 0.2$ & $0.13$ \\
        & Square-Root    & $43.0 \pm 0.2$ & $42.9 \pm 0.2$ & $0.12$ \\
        & Progressive    & $44.1 \pm 0.5$ & $43.9 \pm 0.3$ & $0.22$ \\
        \midrule
        \multirow{4}{*}{100}
        & Uniform        & $39.7 \pm 0.4$ & $39.7 \pm 0.4$ & $0.06$ \\
        & Class-Balanced & $34.0 \pm 0.9$ & $33.8 \pm 0.9$ & $0.17$ \\
        & Square-Root    & $38.1 \pm 0.2$ & $38.0 \pm 0.1$ & $0.11$ \\
        & Progressive    & $40.0 \pm 0.6$ & $39.7 \pm 0.6$ & $0.30$ \\
        \bottomrule
    \end{tabular}
\end{table}

Two consequences matter for the claims we make.

\paragraph{The tail-class finding is protocol-independent.} Repeating the paired comparison of Section~\ref{sec:paired} on final-epoch checkpoints, progressive sampling still beats uniform sampling on tail accuracy for every seed at both imbalance ratios: the per-seed differences are $+2.16$, $+2.39$ and $+2.94$ points at $\rho = 100$ (mean $+2.49$) and $+3.47$, $+1.53$ and $+0.89$ at $\rho = 50$ (mean $+1.96$). Sign agreement remains 3/3 in both cases. Our central result therefore does not rest on the selection protocol.

\paragraph{The overall-accuracy ordering remains unreliable.} Under the final-epoch protocol the strategy ranking is unchanged at $\rho = 10$ and $\rho = 100$, but at $\rho = 50$ progressive and uniform sampling exchange places ($43.9$ versus $44.0$). This reinforces the conclusion of Section~\ref{sec:paired}: the two are not separable in overall accuracy at this sample size, and the apparent lead of progressive sampling in Table~\ref{tab:overall_results} should not be read as a ranking. Class-balanced sampling remains clearly last at every imbalance ratio under both protocols.

\section{Supplementary Figures}
\label{app:supp_figures}
\subsection{Overall Accuracy Comparison}
\label{app:overall_acc_fig}

\begin{figure}[H]
    \centering
    \includegraphics[width=\textwidth]{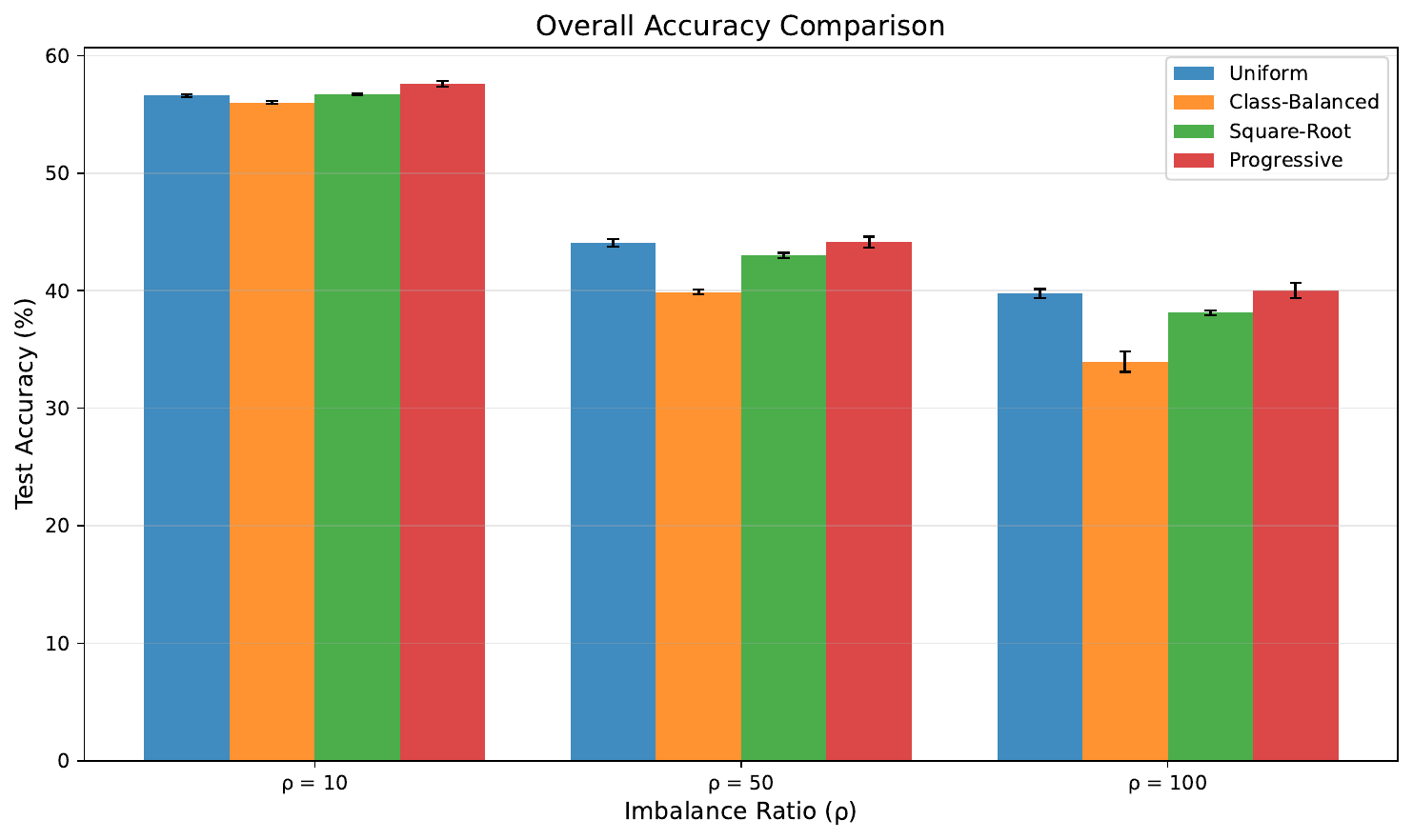}
    \caption{Overall test accuracy comparison across sampling strategies and imbalance ratios. Error bars indicate standard deviation over 3 seeds.}
    \label{fig:overall_comparison}
\end{figure}
\subsection{Head/Medium/Tail Accuracy Bars}
\label{app:hmt_bars_fig}

\begin{figure}[H]
    \centering
    \begin{subfigure}[b]{0.48\textwidth}
        \includegraphics[width=\textwidth]{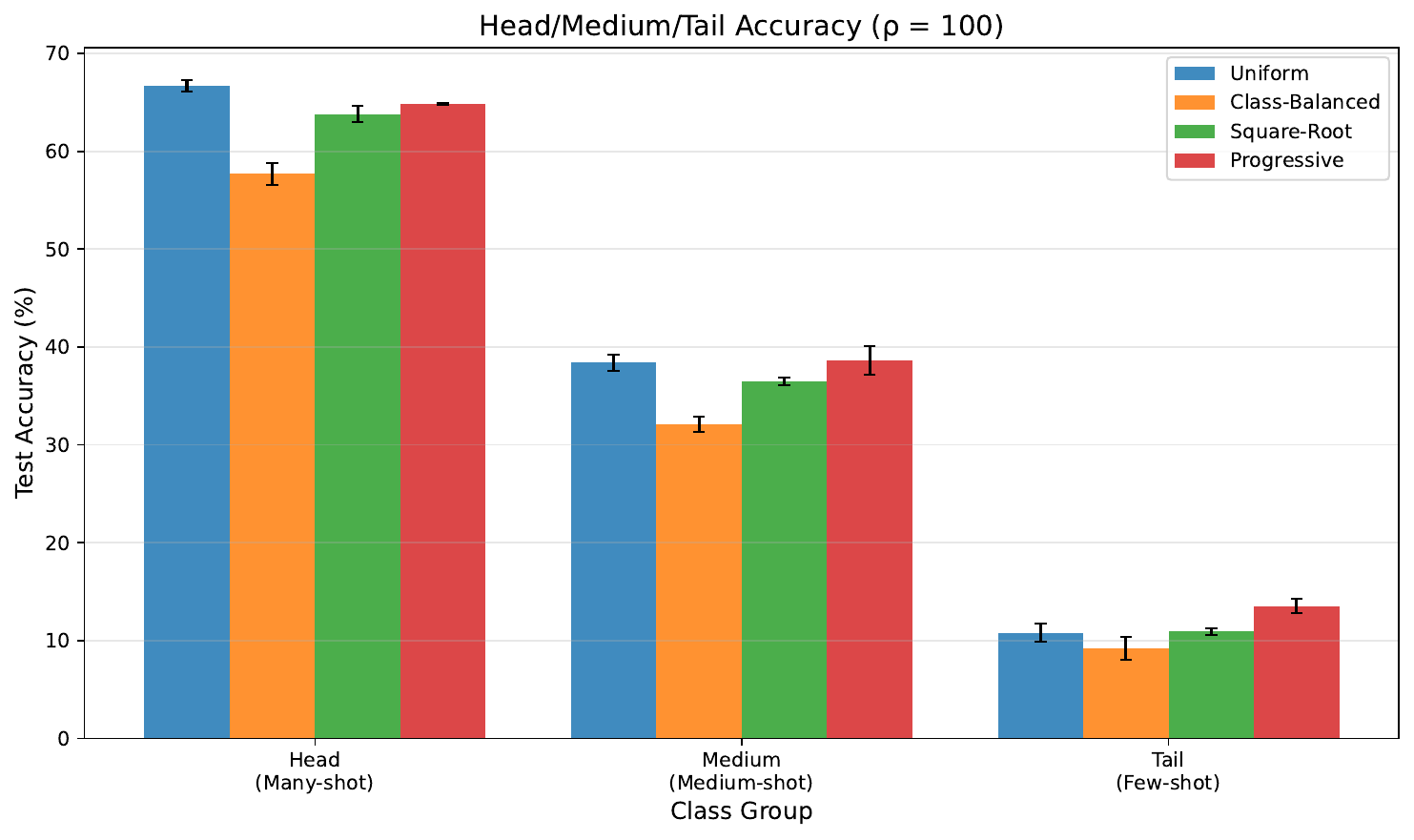}
        \caption{$\rho = 100$}
    \end{subfigure}
    \hfill
    \begin{subfigure}[b]{0.48\textwidth}
        \includegraphics[width=\textwidth]{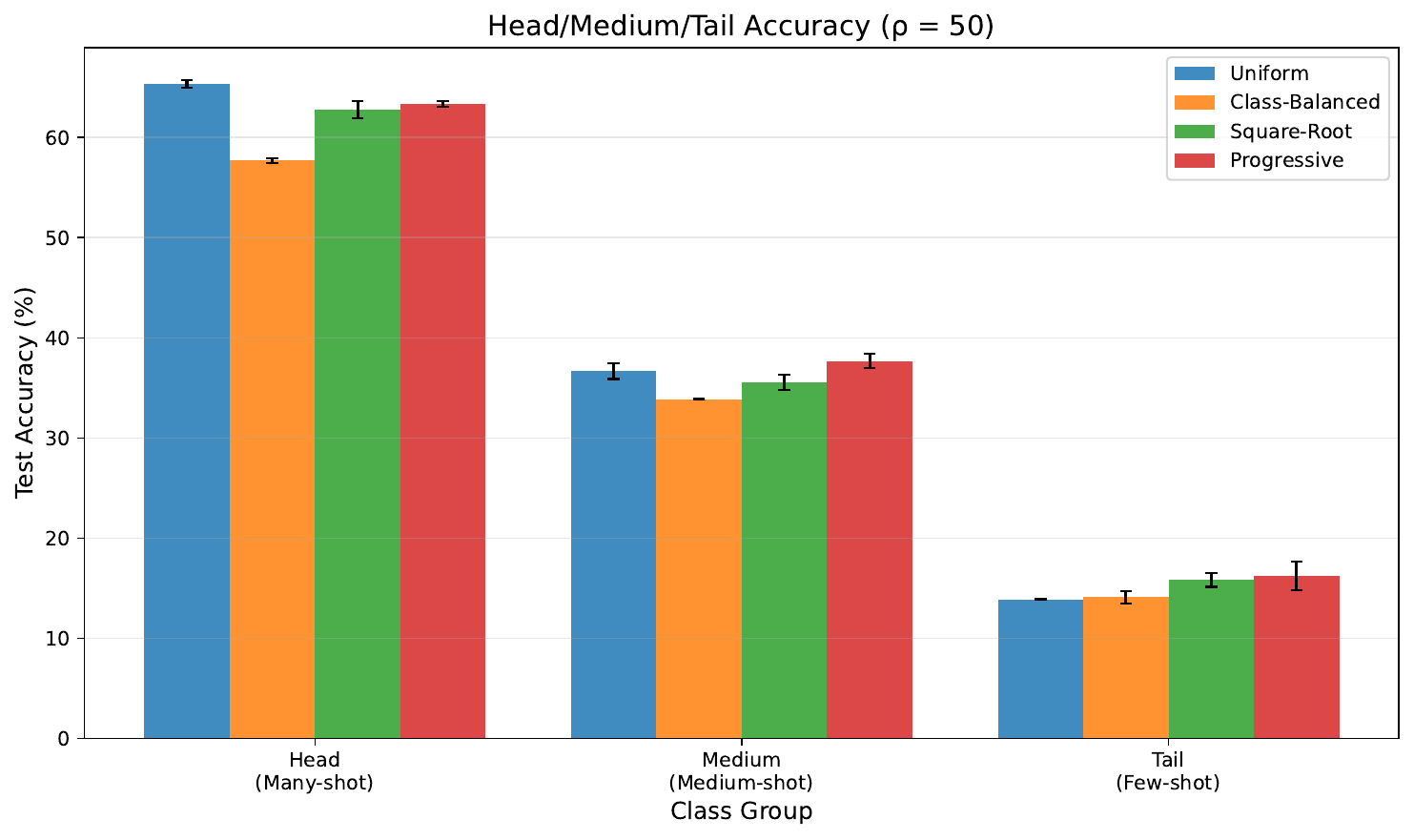}
        \caption{$\rho = 50$}
    \end{subfigure}
    \caption{Head/Medium/Tail accuracy breakdown for $\rho = 100$ and $\rho = 50$. Error bars indicate standard deviation over 3 seeds.}
    \label{fig:hmt_bars}
\end{figure}
\subsection{Learning Rate and Progressive $\lambda$ Schedule}
\label{app:schedule_figures}

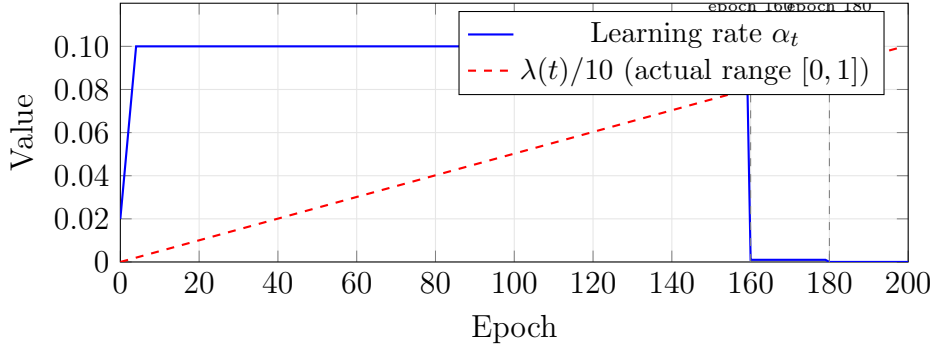
\begin{figure}[H]
    \centering
    \begin{tikzpicture}
        \begin{axis}[
            width=0.75\textwidth,
            height=5cm,
            xlabel={Epoch},
            ylabel={Value},
            xmin=0, xmax=200,
            ymin=0, ymax=0.12,
            legend pos=north east,
            legend style={font=\small},
            grid=major,
            grid style={gray!20},
            ytick={0, 0.02, 0.04, 0.06, 0.08, 0.10},
            yticklabels={$0$, $0.02$, $0.04$, $0.06$, $0.08$, $0.10$},
        ]
        
        \addplot[blue, thick] coordinates {
            (0, 0.02) (1, 0.04) (2, 0.06) (3, 0.08) (4, 0.10)
            (5, 0.10) (159, 0.10) (160, 0.001) (179, 0.001) (180, 0.00001) (200, 0.00001)
        };
        \addlegendentry{Learning rate $\alpha_t$}
        
        \addplot[red, thick, dashed] coordinates {
            (0, 0.0) (20, 0.01005) (40, 0.02010) (60, 0.03015)
            (80, 0.04020) (100, 0.05025) (120, 0.06030)
            (140, 0.07035) (160, 0.08040) (180, 0.09045) (199, 0.10)
        };
        \addlegendentry{$\lambda(t) / 10$ (actual range $[0, 1]$)}
        
        \draw[gray, dashed, thin] (axis cs:160,0) -- (axis cs:160,0.12);
        \draw[gray, dashed, thin] (axis cs:180,0) -- (axis cs:180,0.12);
        \node[font=\tiny, above] at (axis cs:160,0.11) {epoch 160};
        \node[font=\tiny, above] at (axis cs:180,0.11) {epoch 180};
        
        \end{axis}
    \end{tikzpicture}
    \caption{Learning rate schedule (blue) and progressive sampling parameter $\lambda(t)$ (red, dashed; scaled by $\times 0.1$ for visual clarity; actual range is $[0, 1]$). Vertical dashed lines mark the learning rate milestones at epochs 160 and 180.}
    \label{fig:lr_lambda_schedule}
\end{figure}

Figure~\ref{fig:lr_lambda_schedule} illustrates the learning rate schedule and the progressive sampling parameter $\lambda(t)$ over the course of training. The learning rate undergoes linear warmup during the first 5 epochs, remains constant at $0.1$ until epoch 160, then decays by a factor of $0.01$ at epochs 160 and 180. The progressive parameter $\lambda(t) = \left(\frac{t}{T-1}\right)^\gamma$ increases linearly from 0 to 1, reaching $\lambda \approx 0.80$ at the first learning rate milestone.

\subsection{Convergence Curves for $\rho = 50$ and $\rho = 10$}
\label{app:convergence_extra}
\begin{figure}[H]
    \centering
    \includegraphics[width=\textwidth]{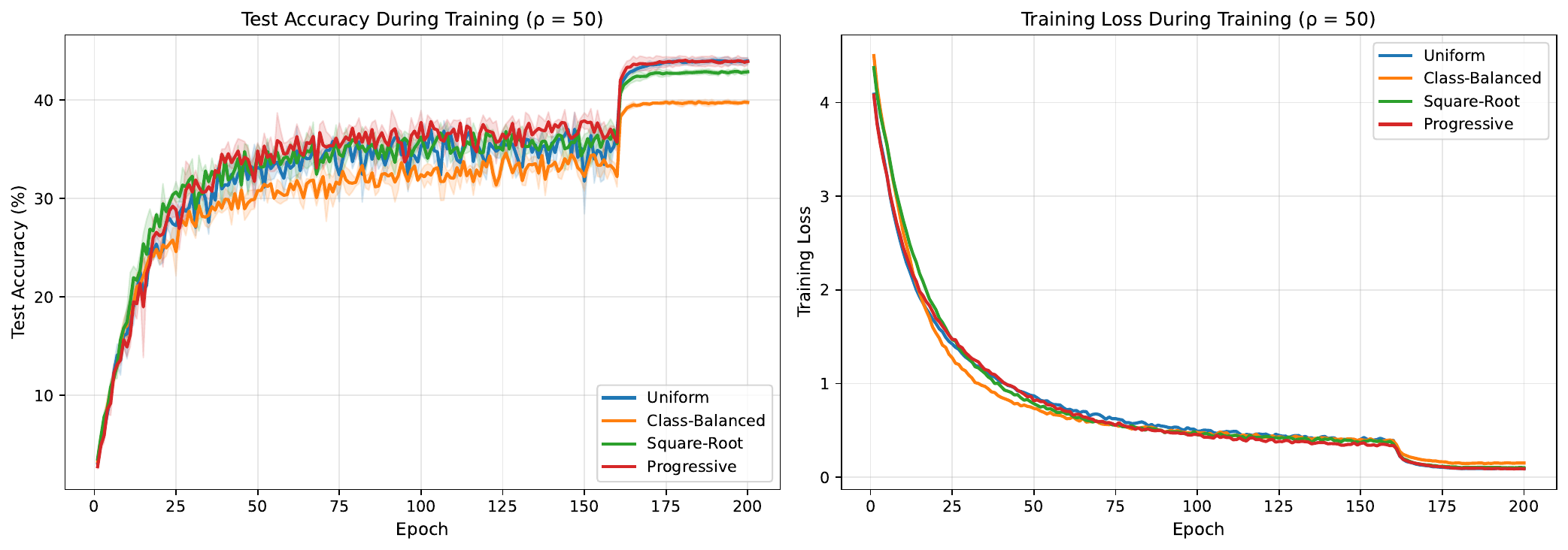}
    \caption{Convergence curves for $\rho = 50$. Left: test accuracy. Right: training loss. Shaded regions indicate $\pm 1$ standard deviation over 3 seeds.}
    \label{fig:convergence_rho50_app}
\end{figure}

\begin{figure}[H]
    \centering
    \includegraphics[width=\textwidth]{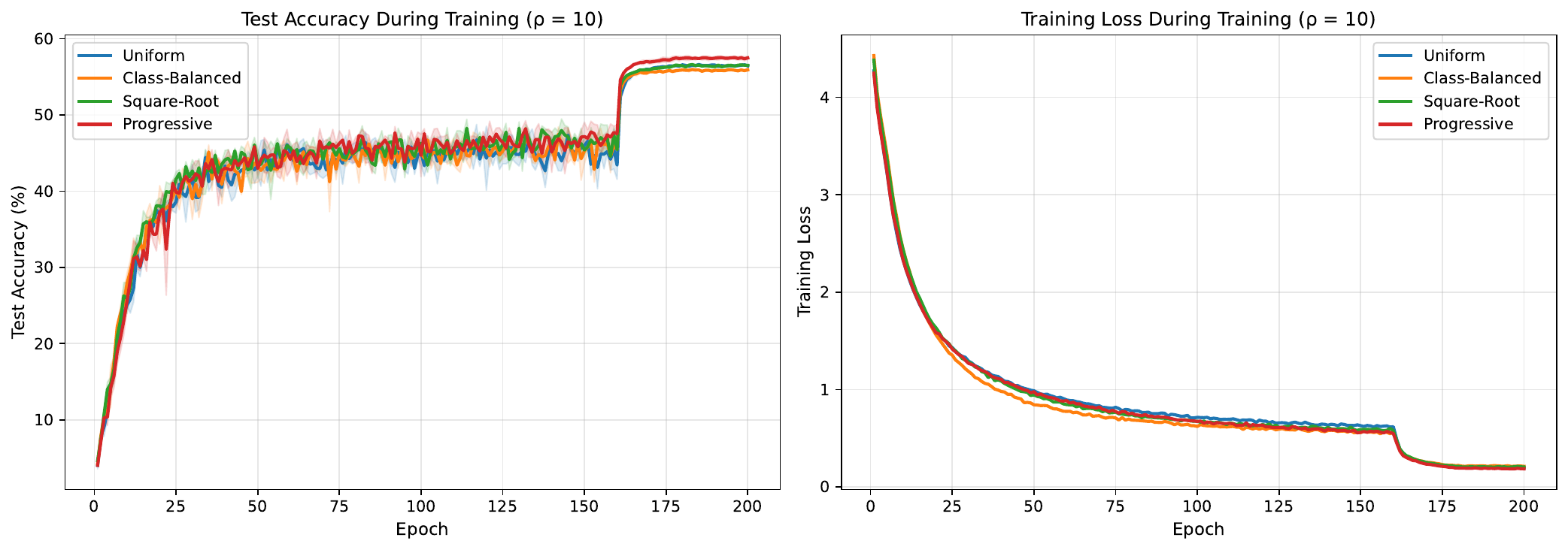}
    \caption{Convergence curves for $\rho = 10$. All four strategies converge to similar final accuracy, reflecting the mild degree of imbalance.}
    \label{fig:convergence_rho10_app}
\end{figure}
\subsection{Per-Group Convergence Curves}
\label{app:per_group_convergence}

\begin{figure}[H]
    \centering
    \includegraphics[width=\textwidth]{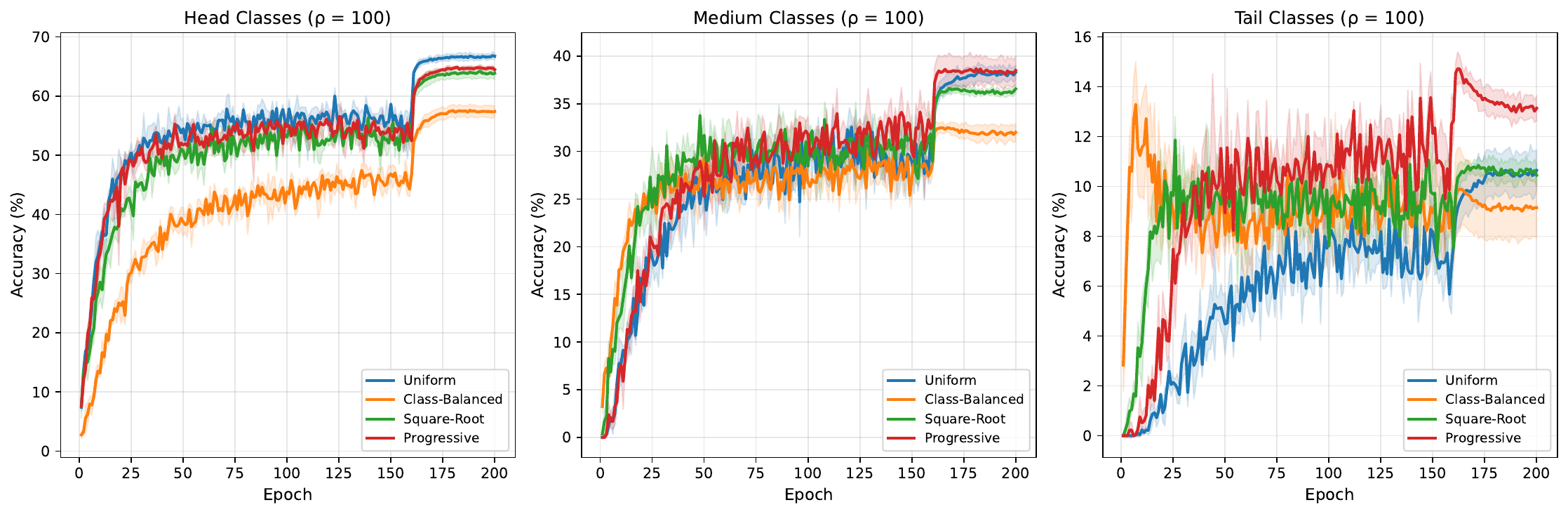}
    \caption{Per-group accuracy during training for $\rho = 100$. Left: head (many-shot). Centre: medium (medium-shot). Right: tail (few-shot). Shaded regions indicate $\pm 1$ standard deviation over 3 seeds.}
    \label{fig:group_curves_rho100_app}
\end{figure}

\begin{figure}[H]
    \centering
    \includegraphics[width=\textwidth]{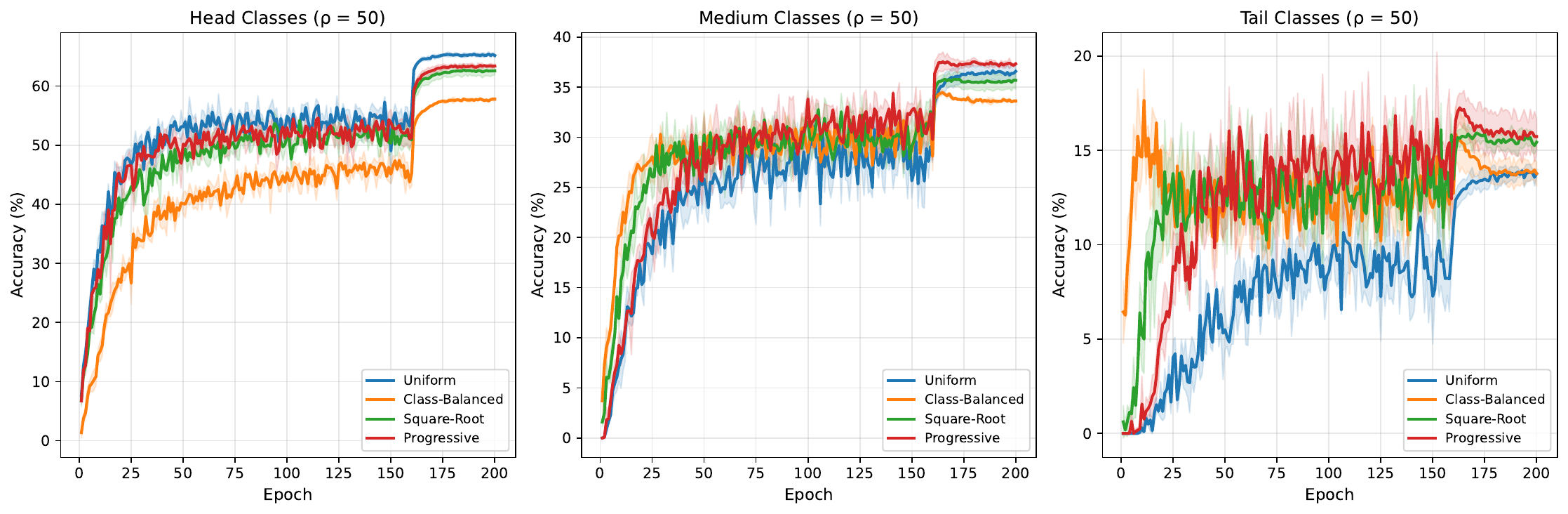}
    \caption{Per-group accuracy during training for $\rho = 50$.}
    \label{fig:group_curves_rho50_app}
\end{figure}

\begin{figure}[H]
    \centering
    \includegraphics[width=\textwidth]{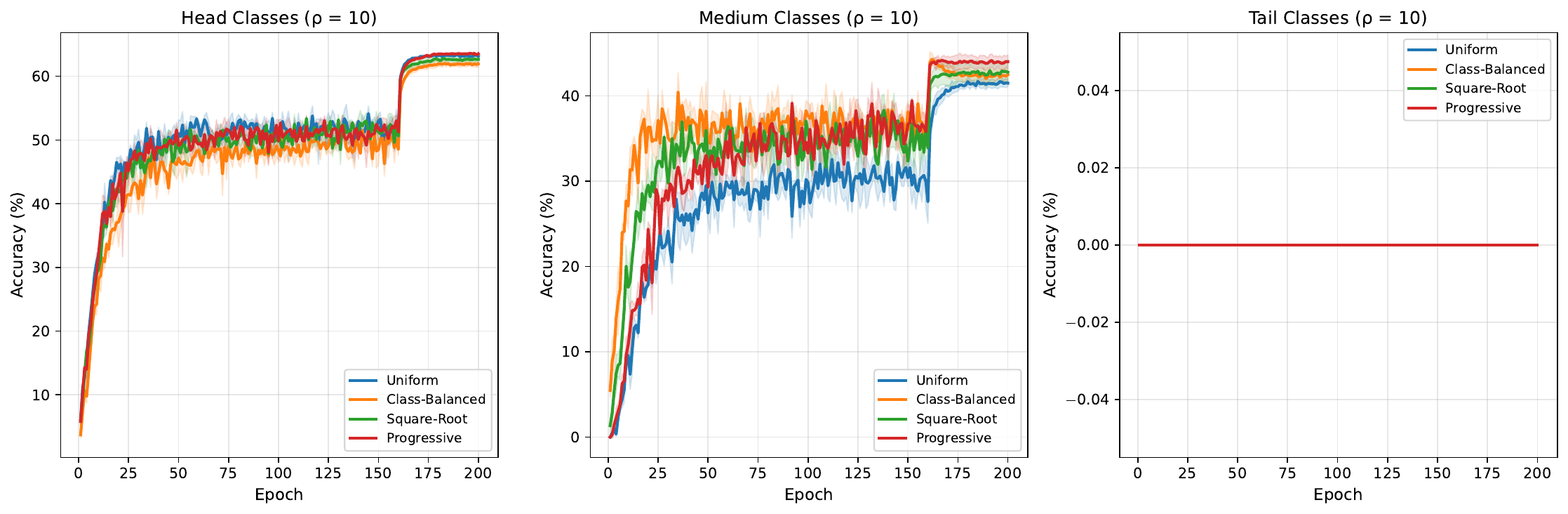}
    \caption{Per-group accuracy during training for $\rho = 10$. Note: the tail group is empty at this imbalance ratio, so the rightmost panel is blank.}
    \label{fig:group_curves_rho10_app}
\end{figure}

Figures~\ref{fig:group_curves_rho100_app}--\ref{fig:group_curves_rho10_app} present the per-group (head, medium, tail) accuracy curves during training for all three imbalance ratios.

\section{Dataset Class Distribution}
\label{app:distribution}

\begin{figure}[H]
    \centering
    \begin{tikzpicture}
        \begin{axis}[
            width=0.85\textwidth,
            height=7cm,
            xlabel={Class index (sorted by frequency)},
            ylabel={Number of training samples},
            xmin=0, xmax=101,
            ymin=0, ymax=550,
            legend pos=north east,
            legend style={font=\small},
            grid=major,
            grid style={gray!20},
        ]
        
        \addplot[blue, thick, mark=none] table[row sep=crcr] {
            x y \\
            1  500 \\
            10 405 \\
            20 321 \\
            30 254 \\
            40 201 \\
            50 159 \\
            60 126 \\
            70 100 \\
            80 79  \\
            90 63  \\
            100 49 \\
        };
        \addlegendentry{$\rho = 10$}
        
        \addplot[orange, thick, mark=none] table[row sep=crcr] {
            x y \\
            1  500 \\
            10 350 \\
            20 235 \\
            30 158 \\
            40 107 \\
            50 72  \\
            60 48  \\
            70 32  \\
            80 22  \\
            90 14  \\
            100 9  \\
        };
        \addlegendentry{$\rho = 50$}
        
        \addplot[red, thick, mark=none] table[row sep=crcr] {
            x y \\
            1  500 \\
            10 328 \\
            20 206 \\
            30 129 \\
            40 81  \\
            50 51  \\
            60 32  \\
            70 20  \\
            80 12  \\
            90 7   \\
            100 5  \\
        };
        \addlegendentry{$\rho = 100$}
        
        \draw[gray, dashed, thin] (axis cs:0,100) -- (axis cs:101,100);
        \node[font=\tiny, right, gray] at (axis cs:75,108) {Head/Medium threshold ($n_k = 100$)};
        
        \draw[gray, dotted, thin] (axis cs:0,20) -- (axis cs:101,20);
        \node[font=\tiny, right, gray] at (axis cs:75,28) {Medium/Tail threshold ($n_k = 20$)};
        
        \end{axis}
    \end{tikzpicture}
    \caption{Training set class distribution for CIFAR-100-LT at $\rho \in \{10, 50, 100\}$. Classes are sorted in descending order of frequency. Horizontal dashed lines indicate the head/medium ($n_k = 100$) and medium/tail ($n_k = 20$) group boundaries.}
    \label{fig:class_dist}
\end{figure}
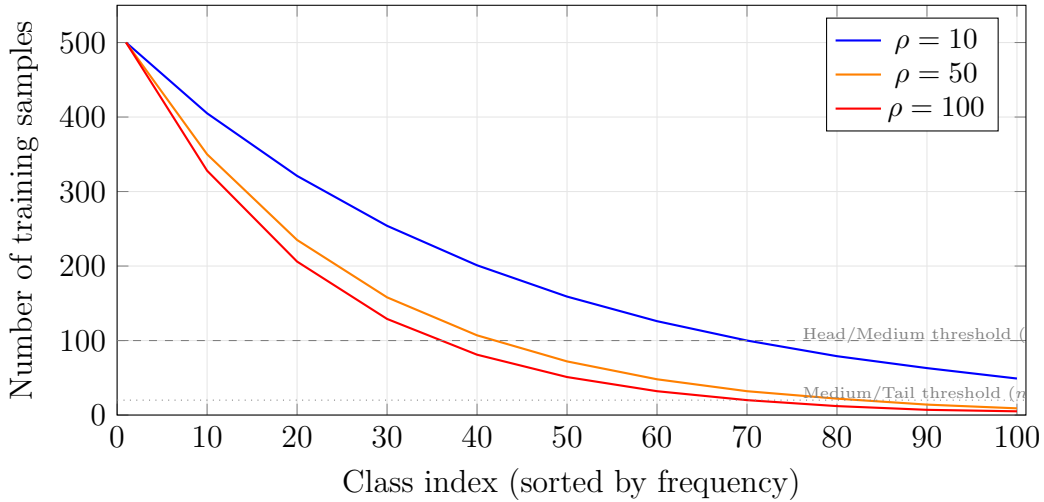

Figure~\ref{fig:class_dist} visualises the number of training samples per class for each imbalance ratio. The classes are sorted by frequency (class index 1 has the most samples, class index 100 the fewest). The exponential decay defined in Equation~\eqref{eq:exponential_decay} produces the characteristic ``long tail'' shape, with the severity controlled by $\rho$.

Table~\ref{tab:group_counts} reports the number of classes assigned to each frequency group for each imbalance ratio.

\begin{table}[H]
    \centering
    \caption{Number of classes in each frequency group for different imbalance ratios.}
    \label{tab:group_counts}
    \begin{tabular}{lccc}
        \toprule
        \textbf{Group (threshold)} & $\boldsymbol{\rho = 10}$ & $\boldsymbol{\rho = 50}$ & $\boldsymbol{\rho = 100}$ \\
        \midrule
        Head ($n_k > 100$)       & 69 & 41 & 35 \\
        Medium ($20 < n_k \leq 100$) & 31 & 40 & 34 \\
        Tail ($n_k \leq 20$)    & 0  & 19 & 31 \\
        \midrule
        Total                    & 100 & 100 & 100 \\
        \bottomrule
    \end{tabular}
\end{table}

\end{document}